\documentclass[12pt,letterpaper]{article}
\usepackage[a4paper, total={7in, 10in}]{geometry}

\usepackage{helvet}
\usepackage{authblk}
\usepackage{hyperref}

\usepackage{amsmath}
\usepackage{amssymb}
\usepackage{graphicx}
\usepackage{booktabs}
\usepackage{multirow}
\usepackage{wrapfig}
\usepackage{pdfpages}

\usepackage[left]{lineno}

\makeatletter
\renewcommand{\maketitle}{\bgroup\setlength{\parindent}{0pt}
\begin{flushleft}
  \textbf{\@title}
  
  \@author
\end{flushleft}\egroup}
\makeatother

\title{Incorporating simulated spatial context information improves the effectiveness of contrastive learning models}
\date{}

\author[1]{Lizhen Zhu}
\author[1,2,3]{James Z. Wang}
\author[4]{Wonseuk Lee}
\author[5]{Brad Wyble}

\affil[1]{Data Science and Artificial Intelligence Area, College of Information Sciences and Technology, The Pennsylvania State University, University Park, PA, USA}
\affil[2]{Human-Computer Interaction Area, College of Information Sciences and Technology, The Pennsylvania State University, University Park, PA, USA}
\affil[3]{Department of Communication and Media, School of Social Sciences and Humanities, Loughborough University, Loughborough, Leicestershire, UK}
\affil[4]{Department of Computer Science and Engineering, The Pennsylvania State University, University Park, PA, USA}
\affil[5]{Department of Psychology, The Pennsylvania State University, University Park, PA, USA}
\affil[*]{Correspondence: bpw10@psu.edu}

\usepackage[super,comma,sort&compress]{natbib}

\begin{document}

\maketitle

\begin{figure*}[ht!]
    \centering
    \includegraphics[width=0.9\textwidth]{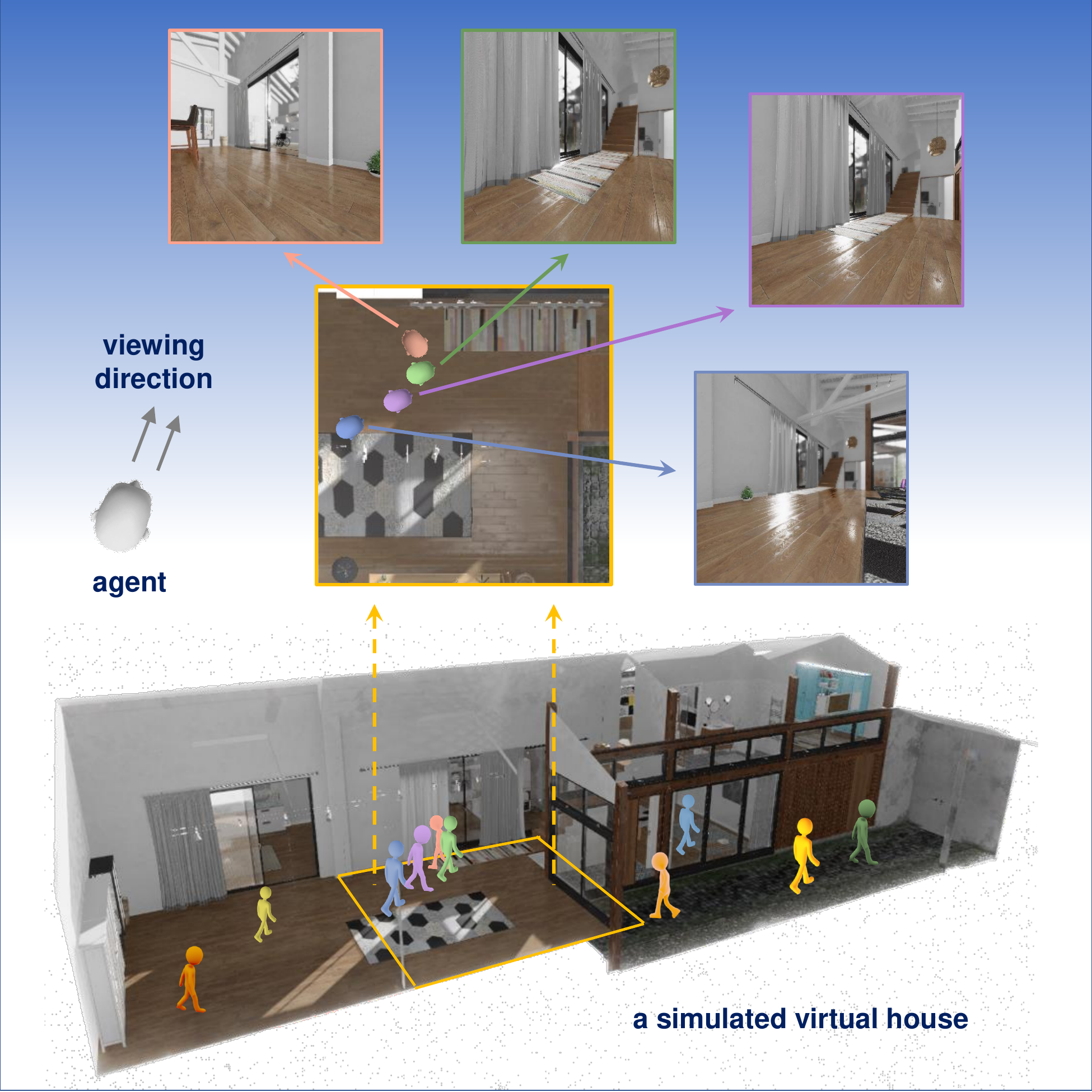}
\end{figure*}


\section*{Summary}
Visual learning often occurs in a specific context, where an agent acquires skills through exploration and tracking of its location in a consistent environment. The historical spatial context of the agent provides a similarity signal for self-supervised contrastive learning. We present a unique approach, termed Environmental Spatial Similarity (ESS), that complements existing contrastive learning methods. Using images from simulated, photorealistic environments as an experimental setting,  we demonstrate that ESS outperforms traditional instance discrimination approaches. Moreover, sampling additional data from the same environment substantially improves accuracy and provides new augmentations. ESS allows remarkable proficiency in room classification and spatial prediction tasks, especially in unfamiliar environments. This learning paradigm has the potential to enable rapid visual learning in agents operating in new environments with unique visual characteristics. Potentially transformative applications span from robotics to space exploration. Our proof of concept demonstrates improved efficiency over methods that rely on extensive, disconnected datasets.

\section*{Keywords}
Contrastive learning, Virtual environment, Developmental psychology, Deep learning, Computer vision, Intelligent agent, Robotics, Childhood learning
\section*{Introduction}

One of the central challenges faced by both artificial and natural cognitive visual systems is the ability to map pixel-level inputs, such as those obtained through eyes or cameras, onto compositional, internal representations that inform decisions, actions, and memory processes. In the recent two decades, significant progress has been made in understanding vision, notably due to the rise of statistical models\citep{li2003automatic,li2008real} and particularly deep neural networks.\citep{krizhevsky2012imagenet} These advances have fostered myriad real-world applications across a wide range of fields, spanning from biomedicine to emotion recognition.\citep{davaasuren2022automated,cai2022deepstroke,luo2020arbee,wang2023emotion}

The process of learning for computational cognitive visual systems often involves the use of vast image datasets that are organized into categories such as specific types of animals or vehicles, or particular concepts such as surface materials,\cite{zheng2016deep} aesthetics,\cite{lu2015rating} or product defects.\cite{yu2022surface}  General-purpose image understanding might use massive datasets, sometimes with billions of images labeled with thousands of discrete linguistic terms,\cite{deng2009imagenet,mahajan2018exploring} but otherwise lack contextual information. For example, two social-media-crawled images labeled as ``French bulldogs'' might both depict different dogs or two views of the same dog. Despite these limitations, these datasets have helped to drive a new generation of deep learning approaches to computer vision, leading to significant improvements in image categorization performance following the release of models such as AlexNet.\cite{krizhevsky2012imagenet} These advances have been achieved through incremental improvements in both the scale and complexity of networks and datasets.

Despite these improvements, deep learning solutions for vision still lack the robustness of human performance, even for the relatively simple task of image recognition. While they perform well on specific target datasets such as ImageNet,\cite{deng2009imagenet}, such models struggle to generalize to other, even highly similar tasks.\cite{recht2019imagenet} Moreover, they lag behind human performance in object classification,\cite{shankar2020evaluating}
and are susceptible to adversarial attacks in ways humans are not.\cite{goodfellow2014generative} Scaling data sets up is not proving an effective remedy for these shortcomings.\citep{mahajan2018exploring} Another drawback of current approaches is that the immense size of large datasets limits the ability to conduct experiments due to restricted access to the images and necessary computing resources, and concerns arise about the environmental toll of the energy used in training.

{\bf Human-inspired contextual learning in computer vision.} To approach this problem we draw inspiration from the nature of human visual learning and how it differs from contemporary computer vision. During their first year or two of life, children are typically extensively exposed to a narrow range of specific visual objects within a highly familiar and constrained context. Many children in modern households spend the first year of their life primarily in one or two buildings, viewing a limited set of spaces, surfaces, faces, and objects from various perspectives and lighting conditions (e.g., sunlight, cloudy light, artificial illumination). Headcam data reveals that only three specific faces comprise the vast majority of face exposure for the average child from Western households in the first year of their life.\cite{jayaraman2015faces} Moreover, children view a comparatively small number of objects, many of which are only seen within a specific context, such as a toaster on a particular kitchen counter with a certain wall texture.  Even the total number of views of the world by a human child is comparatively small compared to the number of images in large data sets. Children typically make around $90$ million visual fixations by the age of two (derived based on an average fixation rate of approximately 1.4/second \cite{papageorgiou2014individual}), which is much smaller than the hundreds of millions or billions of images in the larger datasets. A similar disparity exists for the training of Large Language Models.\cite{frank2023bridging} 

This limited exposure to a narrow range of objects in one context that we see in children would result in poor performance for standard deep learning approaches that typically require balanced exposure to a wide range of objects in different backgrounds to avoid learning skewed statistical relationships. To avoid this problem, large datasets rely on many exemplars of objects on a wide range of backgrounds\cite{tomasev2022pushing} but it is unknown how children learn to effectively parse the visual environment without such diverse visual experiences. To help address this gap, we hypothesize that through the use of environmentally contextualized learning, computer systems can be designed to learn representations that are flexible enough to perform well on generalized tasks such as natural image classification from smaller, less diverse datasets. Our work here provides a step in this direction by showing that including the spatial position of image samples within an environment can measurably improve performance on a task like ImageNet classification relative to an algorithm that uses only instance discrimination for training.

\begin{figure*}[ht!]
    \centering
    \includegraphics[width=0.8\textwidth]{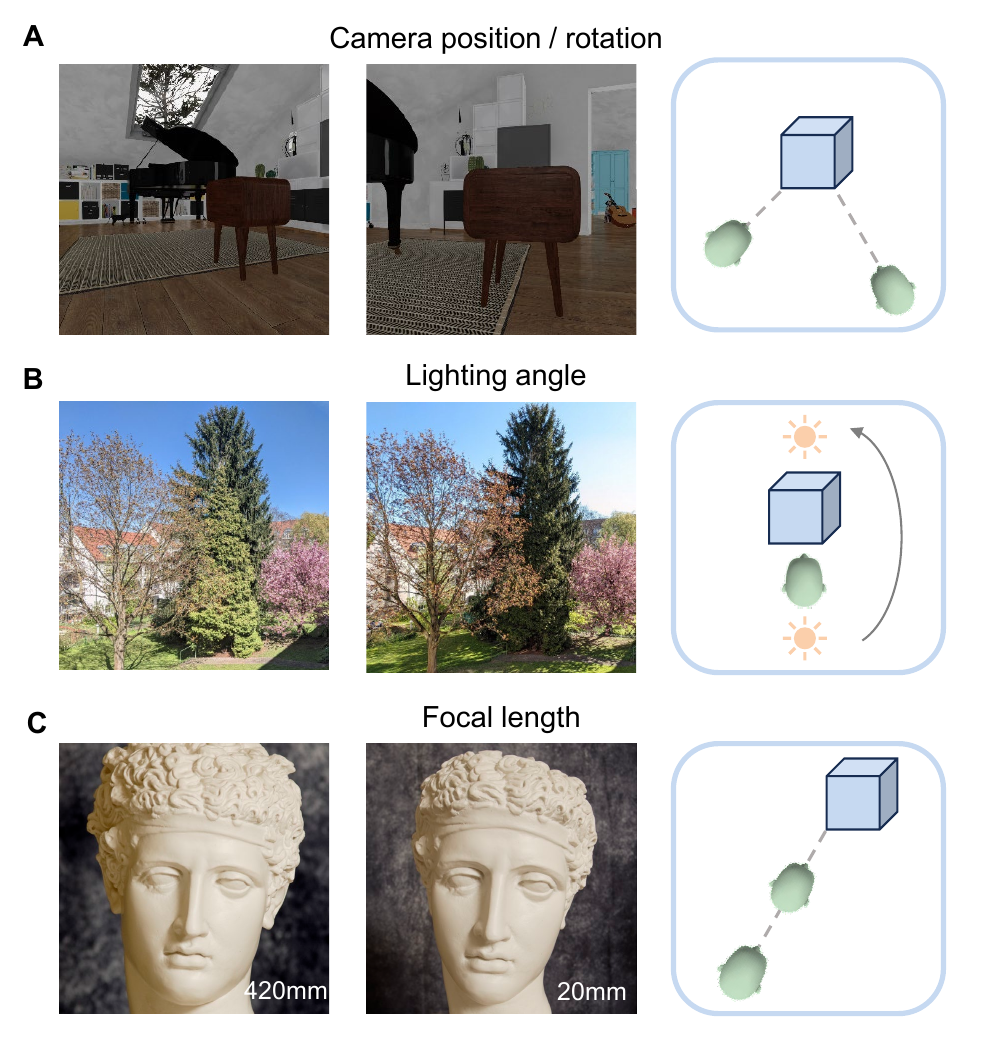}
    \caption{\textbf{The impact of position on the appearance, lighting, and camera distance/focal length of an image}\hfill\break
(A) The perspective of a room can greatly impact its appearance when rendered from different positions in the ThreeDWorld simulated environment.
(B) The natural lighting of a scene can significantly alter its appearance when captured at different times of the day. Photos courtesy of Federico Adolfi.
(C) The head and facial features of a statue may appear differently when captured with different focal lengths. Photos courtesy of James Z. Wang.}\label{fig:imgdif}
\end{figure*}

{\bf Lessons from human visual development.} The field of developmental psychology offers insights into what is missing from contemporary machine vision learning. While viewing the world, children harness a wealth of environmental information about how their bodies deliberately sample information through controlled orientation of their senses and their interactions with the world.\cite{ballard1997deictic, smith2005cognition,campos2000travel} Inspired by these findings, we take an interdisciplinary step by introducing a new learning approach to self-supervised contrastive learning in which the environment is considered as the data source. This approach allows us to repeatedly sample the same objects in the same rooms from slightly varying positions using a notional agent that occupies a specific location at each time point. For example, while a house has a limited set of locations and objects, the number of possible visual patterns that can be experienced within it is vast given the ability to move such an agent around, to experience varying lighting conditions over time, and to vary physical properties of the sensors such as focal depth. Figure~\ref{fig:imgdif} illustrates such visual differences.

In humans, this mechanism may emerge early in the developmental process, perhaps even before a child begins to move independently (i.e., self-locomotion), if they are passively moved and track self-motion through sources such as optic flow, vestibular input, and other senses. This kind of visual learning precedes and enables higher-order learning mechanisms that infer properties about labeled categories,\cite{yu2013joint, pereira2014bottom} causal interactions,\cite{gopnik2004mechanisms} and physical reasoning.\cite{spelke1990principles,ullman2018learning,walker2017effects}

{\bf Improving self-supervised learning.} Self-supervised learning approximates some aspects of early human visual experience by learning visual patterns from unlabeled images. One such algorithm, called contrastive learning,\cite{chen2020mocov2,grill2020bootstrap,chen2020simple} trains networks to detect when two images are algorithmically derived augmentations of a base image (i.e. positive pairs). However, this approach lacks the ability to represent real-world similarity in the training process. Two source images from nearly identical views of an object would be treated as completely distinct by this approach since they are different instances.  On the other hand, human visual learning is thought to exploit the similarity between proximal samples within the environment to develop a smooth latent representation that connects different views of the same objects.\citep{zhuang2021unsupervised} Such similar images are a natural byproduct of perception by any agent that traverses an environment in which objects persist over time, thereby providing a variety of changes in perspective, lighting conditions, and so on. The information that can be extracted from sequential samples by these agents is much richer than what can be gained through instance discrimination alone.  

This aspect of environmentally driven learning transforms the statistical consistency of the world, which might be seen as a disadvantage in some traditional deep learning approaches, into a valuable signal for understanding the physical properties of how light and materials interact for arrangements of objects and surfaces in a visually rich environment, as guided by information about location. This approach is inspired by embodied perspectives on human perception\cite{gibson1966senses} and learning.\cite{campos2000travel,anderson2013role} While a wholly embodied approach would have agents actively engage with their surroundings and learning co-occurs with behavior, our method conducts after the agent has sampled a large set of images. In this approach, positive pairs reflect variations due to both typical augmentations and small shifts in viewing position. Thereby we use the relative positions of the agent at the time two given images were sampled as a proxy of their image similarity. The mechanism we envision does not rely on externally derived labels or even the notion of what objects are. In a cognitive framework, this kind of learning serves as a foundation for subsequent learning at which point the ability to perceive the significance of verbal labels begins to influence visual learning.\cite{pereira2013recognition}

{\bf Our environmental spatial similarity approach.} Our proposed algorithm demonstrates improved efficiency in learning how to visually categorize objects when compared to an existing contrastive learning method. We define increased efficiency as improving accuracy on a downstream ImageNet task while keeping the size of the model, dataset volume, training epochs, augmentations, and downstream task fixed. Our approach involves adjusting the Momentum Contrast (MoCo) algorithm\cite{chen2020mocov2} to leverage spatial context information obtained through simulated images collected in a single environment to determine which images from a randomly sampled dictionary are positive pairs. In MoCo, a positive pair is two augmentations derived from the same source image. In our proposed approach, a positive pair is two images that were proximal in spatial and rotational coordinates. For each key image, there could exist more than one positive pair. We term this approach as Environmental Spatial Similarity with Multi-Binary positive pairs (ESS-MB).
We demonstrate across a variety of conditions that the training process using spatial context to mark positive pairs is more efficient than the same-instance discrimination found in MoCo V2. We further extended the binary representation of similarity to a continuous one to assign differentiated weights to positive pairs, called the multi-weighted version (ESS-MW), resulting in a further modest enhancement in the downstream performance.

We highlight five convergent findings that support the effectiveness of this approach. First, by examining various spatial similarity thresholds, we identify that there is a point of peak performance. Using such a threshold, our approach's pretraining on a set of 102,197 (abbreviated as ``100K'') images collected during one traversal of an environment leads to enhanced downstream accuracy in an ImageNet classification task compared to the MoCo model pretrained on the same image set, with a further improvement using a loss function that is weighted by spatial proximity. This approach is complementary to other contrastive learning models. Second, our approach's superior performance generalizes to both a smaller dataset collected from the same environment and one from a different environment. Thirdly, by accumulating more images of similar views within the same environment, we observe enhanced accuracy, even with the same total amount of training. Moreover, we explore a new form of augmentation afforded by ray tracing with varying light sources and multiple downstream tasks. Lastly, the model with our approach outperforms the MoCo model on room classification and spatial localization tasks, especially in unseen environments. All critical comparisons in our experiments were conducted thrice to offer a confident accuracy range, factoring in the standard error.

\section*{Results}
\subsection*{Simulated datasets provide a source of spatial similarity}

To create a dataset that exhibits environmental consistency, we used a simulation approach that leverages state-of-the-art ray tracing within the Unity framework. Simulations provide us and other researchers the agility to experiment, allowing testing of the effect of highly specific, parametric variations in the image set--something not feasible with real-world image sets.  Building on the ThreeDWorld platform,\citep{gan2020threedworld} we simulated an agent moving through a fully furnished, detailed house and apartment, capturing images at closely spaced intervals. In this environment, ray tracing was used to simulate the transmission of light rays from virtual sources, which bounce and scatter to create realistic perspective, reflections, shadows, and material properties such as glossiness that mimic the appearance of real gloss in human psychophysics.\citep{storrs2021diverse}

The Archviz House (referred to as ``House'') and the Apartment (referred to as ``Apt'') are both simulated building interiors provided by the ThreeDWorld platform. Each is furnished with a set of objects (e.g., furniture, laptop, and cup). The House was enhanced with an additional set of 48 objects, whereas the Apt was enhanced with 101 objects, all of which were sourced from a library of 3D objects using a JSON file.

\begin{figure*}[ht!]
    \centering
    \includegraphics[width=\textwidth]{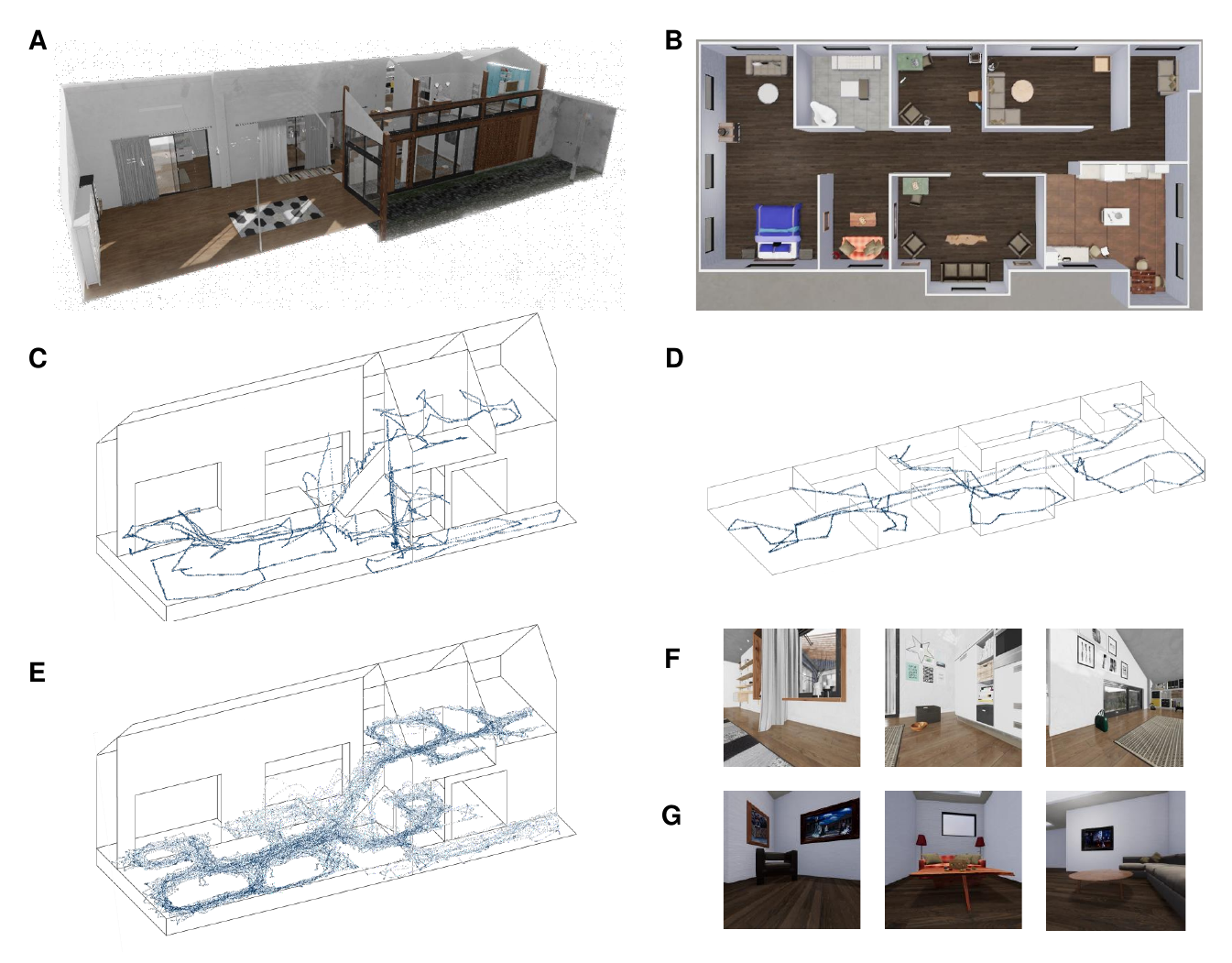}
    \caption{\textbf{The simulated environments and the trajectories used by the embodied agent to generate the datasets}\hfill\break
    (A) The Archviz House. (B) The Apartment. (C-E) The trajectories for House14K, Apt14K, and House100K, respectively. (F-G) Three example images from the House and Apt environments, respectively. During training, random batches were sampled from these trajectories. Images were considered similar if they were spatially close to each other. }
\label{fig:paths}
\end{figure*}

We generated three basic datasets: House14K, House100K, and Apt14K, where the numbers 14K and 100K refer to the approximate number of samples. These datasets were collected under the default lighting condition of ThreeDWorld. Every sample is a $224\times 224$ egocentric image captured by the avatar, accompanied by its respective position and rotation. These samples were generated from pre-recorded avatar trajectories created by a human user navigating the buildings via keyboard controls. Figure~\ref{fig:paths} shows the two simulated environments, the trajectories for all three datasets, and some example images captured within both settings. Within the House environment, we also varied the simulated lighting conditions of simulation to generate House100KLighting and House14KLighting datasets as described in the experimental procedures. 

\subsection*{ESS has superiority over instance discrimination}
To investigate whether our ESS approach improves visual learning, we conducted a study comparing contrastive learning models based on our approach with a self-supervised technique using the identical training set. Our approach modifies the MoCo V2 algorithm by \citeauthor{chen2020mocov2} \cite{chen2020mocov2} Because instance discrimination can only learn if two images are different augmentations of the same image, it overlooks the degree of similarity or difference between distinct images. For the ESS-MB approach, we find similar images in the dictionary for each key image based on the agent's position and rotation and record them as positive pairs. Each positive pair contributes equally to the calculation of the loss function. In the ESS-MW approach, each positive pair is given a weight for loss calculation based on the position and rotation difference between the two images.

We compare our ESS-MB with MoCo V2\cite{chen2020mocov2} when trained on our simulated datasets, specifically the House100K, where images selected for training are randomized in sequence. Unless otherwise specified, ESS-MB represents a variant of MoCo V2 that incorporates ESS-MB. 

Pretext training for the baseline MoCo V2 model\cite{chen2020mocov2} used the same House100K dataset, with the same dictionary size, augmentation techniques, epoch count, batch size, and downstream ImageNet task. All simulations were executed thrice on four NVIDIA RTX A6000 GPUs, with average results and standard error subsequently computed. 

With ESS-MB, the thresholds for distance and rotation similarity serve as adjustable parameters, fine-tuning the spatiotemporal boundaries of environmental consistency. At extremely narrow thresholds (e.g., 0.001 meters or degrees), ESS-MB closely mirrors the instance discrimination as used in MoCo. As presented in Table~\ref{tab:main_result}, the threshold of 0.8 meters and 12 degrees yielded the best downstream performance, with a classification accuracy of $18.05\%$ and degraded performance with both higher and lower threshold values. To gauge the relative importance of both rotation and position, we retrained the model with these variables omitted from the threshold and found that the exclusion of either variable caused a comparable dip in accuracy. For subsequent experiments, the thresholds of 0.8 meters and 12 degrees were retained for the ESS-MB model trained on House100K. A further improvement in accuracy was obtained by introducing a modest quantity of ImageNet-style training images to the pretext training, after which the downstream accuracy was $23.36\%$. More details are provided in Experimental Procedures.

\begin{table*}[ht!]
\caption{\textbf{Comparison between the baseline and ESS on House environment with modified thresholds}\hfill\break ``Threshold'' indicates that if a sample's position and rotation difference relative to the key samples is below (x meters, y degrees), it is designated a positive sample. The column ``Pos. Pairs'' shows the average number of positive samples in the dictionary for each threshold. N/A indicates that one or more thresholds were omitted from the similarity metric. ``House14KLong'' indicates that the number of pretext training epochs was increased to equate total training between the 14K and 100K datasets. $\uparrow$ denotes that higher values of this term are preferable. $\downarrow$ denotes that lower values of this term are more favorable. Numbers after $\pm$ represent the standard error of the mean, rounded to a minimum of .01. Numbers in bold highlight the best downstream classification results on each dataset.}
\vskip 0.05in
\scalebox{0.8}{
    \begin{tabular}{l l l l c c c c}
    \toprule
        Pretext & \multicolumn{2}{c}{Training Stage} & Pos. & {Pretext Task} & \multicolumn{3}{c}{Downstream ImageNet Classification} \\
         Dataset & Model & Threshold & Pairs  &Training loss $\downarrow$ & Training loss $\downarrow$ & Test loss $\downarrow$ &Test accuracy ($\%$) $\uparrow$ \\
        \hline
         \multirow{6}{*}{House100K}& Baseline&N/A &1& $4.43\pm 0.02$ &$4.71\pm 0.03$ & $4.72\pm 0.03$ &$17.36 \pm 0.36$\\
         &ESS-MB&(0.4,6) &1.3& $3.78\pm 0.01$ &$4.68\pm 0.03$&$4.71\pm 0.03$& $17.56\pm 0.34$\\
         &ESS-MB&(0.8,12) &6.3& $4.00\pm 0.00$ &$4.67\pm 0.00$&$4.75\pm 0.01$& $\textbf{18.05}\pm 0.04$\\
         &ESS-MB&(1.6,24)&29.3& $4.57\pm 0.00$ &$4.86\pm 0.03$&$4.97\pm 0.02$& $16.92\pm 0.19$\\
         &ESS-MB&(0.8,N/A) &60.5 & $4.81\pm 0.00$ &$5.02\pm 0.04$&$5.14\pm 0.03$& $15.92\pm 0.15$\\
         &ESS-MB&(N/A,12) &292.4 & $5.97\pm 0.00$ &$4.94\pm 0.01$&$4.88\pm 0.01$& $15.55\pm 0.15$\\
         \hline
         \multirow{2}{*}{House14KLong}& Baseline & N/A & 1 & $3.72\pm 0.00$ & $5.18\pm 0.01$ & $5.17\pm 0.01$ & $12.44\pm 0.12$\\
         & ESS-MB &(0.5,7.5) &6.6& $3.87\pm 0.00$ &$5.09\pm 0.00$&$5.11\pm 0.01$& $\textbf{13.44}\pm 0.08$\\
         \hline
         \multirow{4}{*}{House14K}&Baseline&N/A&1& $5.19\pm 0.09$ &$6.79\pm 0.52 $& $6.92\pm 0.55$ &$9.46 \pm 0.79$\\
         &ESS-MB &(0.25,3.75)&2.6& $5.23\pm 0.02$ &$5.89\pm 0.09$&$5.96\pm 0.10$& $11.09\pm 0.29$\\
         &ESS-MB &(0.5,7.5) &6.6 & $5.19\pm 0.01$ &$5.59\pm 0.01$&$5.63\pm 0.01$& $\textbf{11.61}\pm 0.16$\\
         &ESS-MB &(1.0,15) &20.6 & $5.41\pm 0.06$ &$5.62\pm 0.05$&$5.66\pm 0.04$& $11.07\pm 0.10$\\
         \hline
         \multirow{4}{*}{Apt14K}& Baseline &N/A&1&$5.29\pm 0.10$ &$17.37\pm 10.45$ & $18.30\pm 11.61$ & $5.54\pm 0.88$ \\
         &ESS-MB &(0.3,4.5)&2.4& $5.41\pm 0.04$ &$9.28\pm 1.78$&$9.31\pm 1.69$& $6.87\pm 0.48$\\
         &ESS-MB &(0.6,9) &6.7& $5.40\pm 0.02$ &$6.55\pm 0.06$&$6.55\pm 0.02$& $\textbf{8.47}\pm 0.09$\\
         &ESS-MB &(1.2,18) &21.2 & $5.58\pm 0.07$ &$6.46\pm 0.11$&$6.54\pm 0.15$& $8.28\pm 0.29$\\
    \bottomrule
    \end{tabular}}
\label{tab:main_result}
\end{table*}

Expecting the model to learn more effective information from the continuous similarity representation, we developed the ESS-MW approach which added a weight to each positive pair in the loss function. The weight increases as the position and rotation of two samples become closer. As shown in Table~\ref{tab:weighted}, with the thresholds of (0.8,12), (0.4,6), and (1.6,24), ESS-MW improved the test accuracy by $0.39\%$, $0.47\%$, and $0.69\%$, respectively, compared to ESS-MB.

\begin{table*}[ht!]
\caption{\textbf{Result of ESS-MW with various thresholds on House100K environment}\hfill\break }
\vskip 0.05in
\centering
\scalebox{0.8}{
    \begin{tabular}{l p{0.01cm} c p{0.01cm} c c c}
    \toprule
        Training Stage && {Pretext Task} && \multicolumn{3}{c}{Downstream ImageNet Classification} \\
         Threshold &  &Training loss $\downarrow$ &  & Training loss $\downarrow$ & Test loss $\downarrow$ &Test accuracy ($\%$) $\uparrow$ \\
        \hline
         (0.8,12)& & $3.92\pm 0.003$ &  &$4.62\pm 0.001$&$4.67\pm 0.004$& $18.39\pm 0.082$\\
         (0.4,6)& & $3.77\pm 0.004$ &  &$4.98\pm 0.326$&$4.67\pm 0.024$& $18.03\pm 0.114$\\
         (1.6,24)& & $4.26\pm 0.003$ &  &$4.72\pm 0.011$&$4.82\pm 0.013$& $17.61\pm 0.135$\\
    \bottomrule
    \end{tabular}}
    \label{tab:weighted}
\end{table*}

These downstream accuracy scores compare favorably to results from the MoCo model trained on the same dataset, with average scores from the baseline model trailing behind the ESS-MB average by a margin exceeding standard errors. These results suggest that spatial similarity context facilitates learning from the pretext task on simulated images in a way that translates to the superior classification of ImageNet images that the representational backbone model has never been trained on.

We tested whether these results generalize to smaller datasets within the same environment using just 14K images and for a different environment entirely. Specifically, we used the House14K and Apt14K datasets. For the House100K dataset, the most effective threshold settings yielded an average of 6.3 positive pairs in the dictionary for each image. To bring the average number of positive pairs to be around $6.5$ for the 14K datasets, the best thresholds were found to be 0.5 meters and 7.5 degrees for House14K, and 0.6 meters and 9 degrees for Apt14K. With these thresholds, ESS-MB also outperformed MoCo with downstream accuracies of $11.61\%$ and $8.47\%$, compared to  $9.46\%$ and $5.5\%$ for the baseline MoCo models trained on the same datasets.

\subsection*{Richer exploration of an environment improves learning}
Training with the House100K dataset produces a substantially higher accuracy on ImageNet classification for both models, even though both the House14K and House100K datasets contain images from the same rooms. The improvement in performance might stem from the larger number of training steps involved with the House100K dataset. To control for this factor, we trained the ESS-MB model on the House14K dataset for 1,428 epochs, which is equivalent to the total number of training steps in the House100K dataset over 200 epochs. Nevertheless, even when equating training steps, the House14K dataset yielded lower downstream accuracy than the House100K dataset, by a margin of $4.61\%$, as shown in Table~\ref{tab:main_result}. These results support the hypothesis that a more extensive exploration within a single environment can lead to improved performance, both in terms of distinguishing features within that environment and in the supervised classification of real-world images.

\subsection*{ESS is complementary to other contrastive learning approaches}
Our ESS approach could be applied to most constrastive learning model to improve their performance. We further implemented our ESS-MB approach on SimCLR,\cite{chen2020simple} Decoupled Contrastive Learning (DCL),\cite{yeh2022decoupled} and the Contrastive Learning with Stronger Augmentations (CLSA)\cite{wang2022contrastive} on House100K to determine if our approach improves performance for these algorithms. Note that the Nearest-Neighbor Contrastive Learning of visual Representations (NNCLR)\cite{dwibedi2021little} uses a different way to define the positive pairs, so that we could not implement ESS-MB on NNCLR. Instead, we compared ESS-MB on MoCo with NNCLR using the ResNet-18\cite{he2016deep} backbone. In addition, we implemented ESS-MB on MoCo V3\cite{Chen_2021_ICCV} with the Vision Transformer (ViT)\cite{dosovitskiy2021image} backbone. For more details, please refer to the Supplemental Information.

As shown in Table~\ref{tab:otherCLmodels}, on all 5 models, our approach outperforms the original one. For SimCLR and MoCo V3, both models use batch-wise contrast. A total batch size of 1024 of 4 GPUs limits the number of positive and negative pairs that can be obtained. With the same threshold, there are only 1.6 positive pairs for each image on average, thus leading to limited influence on the model performance. For NNCLR, although ESS-MB and NNCLR reported closely matched losses of 3.39 and 3.89 in the pretext task, our model achieved an accuracy of $7.96\%$ on the downstream ImageNet classification task--a marked improvement of $4.41\%$ over NNCLR. The large downstream training loss of NNCLR is related to the implementation of Lightly.\cite{susmelj2020lightly}

\begin{table*}[ht!]
\caption{\textbf{Comparison of the ESS-MB with various contrastive learning models trained on House100K \hfill\break } The $\checkmark$ means ESS-MB is implemented on a specified contrastive learning model. We compare NNCLR with ESS-MB on MoCo V2, as NNCLR's different definition of positive pairs complicates the direct application of ESS-MB on NNCLR. Numbers in bold highlight the better downstream classification results for each model type.}
\vskip 0.05in
\centering
\scalebox{0.8}{    
    \begin{tabular}{l c c p{0.2cm} l p{0.2cm} c c c}
    \toprule
        \multirow{2}{*}{CL Model} &\multirow{2}{*}{ESS-MB} & \multirow{2}{*}{CL backbone} & & {Pretext Task} && \multicolumn{3}{c}{Downstream ImageNet Classification} \\
        & & & & Training loss $\downarrow$ & & Training loss  $\downarrow$& Test loss $\downarrow$ &Test accuracy ($\%$) $\uparrow$\\
        \hline
        SimCLR&&ResNet-50 & &$0.15\pm 0.01$& &$4.88\pm 0.01$&$4.84\pm 0.01$&$16.81\pm 0.05$\\
        SimCLR&\checkmark&ResNet-50 & &$0.59\pm 0.01$ & &$4.70\pm 0.01$ &$4.79\pm 0.01$ & $\textbf{17.71}\pm 0.13$\\
        \hline
        DCL&&ResNet-50& &$3.75\pm 0.06$&  &$4.67\pm 0.01$&$4.70\pm0.01$&$17.62\pm 0.11$ \\
        DCL&\checkmark& ResNet-50& & $3.86\pm 0.00$&&$4.66\pm 0.01$&$4.69\pm 0.02$&$\textbf{18.15}\pm 0.10$\\
        \hline
        CLSA&&ResNet-50& & $11.44\pm 0.00$&&$4.16\pm 0.03$&$4.06\pm0.03$& $24.77\pm 0.33$\\
        CLSA&\checkmark& ResNet-50& &$11.23\pm 0.00$&&$3.89\pm 0.01$&$3.83\pm0.01$&$
        \textbf{27.77}\pm 0.22$\\
        \hline
        NNCLR&&ResNet-18& &$3.39\pm0.24$&  &$1555\pm8.26 $&$7.03\pm 0.15 $&$3.55\pm 0.03$\\
        MoCo V2&\checkmark& ResNet-18& & $3.89\pm 0.10$& &$5.75\pm 0.01$&$5.71\pm 0.01$&$\textbf{7.96}\pm 0.08$\\
        \hline
        MoCo V3&&ViT& & $1.87\pm 0.01$ & &$4.58\pm 0.02$&$4.47\pm 0.02$&$19.27\pm 0.21$ \\
        MoCo V3&\checkmark& ViT& &$2.11\pm 0.00$ & &$4.57\pm 0.01$&$4.46\pm 0.01$&$\textbf{19.84}\pm0.13$\\   
    \bottomrule
    \end{tabular}}
    \label{tab:otherCLmodels}
\end{table*}

\subsection*{Simulated lighting is a complementary augmentation}
In traditional contrastive learning, augmentations such as random cropping, Gaussian blur, and color jittering are used to train the model to be invariant to minor image variations. However, these techniques fail to capture realistic variation in lighting due to changes in the illuminant which happen in real-world viewing conditions, particularly when observing the same location at different times of the day. To evaluate whether simulated images from different lighting conditions could serve as a complementary source of augmentations, we developed the House100KLighting dataset that uses nine different lighting settings. We conducted three experiments to investigate the impact of lighting-based augmentation on classification results. First, we removed the traditional augmentations from ESS-MB. Second, we excluded standard augmentations and trained ESS-MB with House100KLighting instead of House100K. To make the number of training samples the same, for each image, we randomly selected one of the nine lighting conditions shown in Figure~\ref{fig:light} from the dataset. Third, we trained the ESS-MB using both House100KLighting and the standard augmentations. As shown in Table~\ref{tab:augmentation_result}, the pretext task losses remain unaffected. There was a decline of $8.38\%$ in downstream accuracy when augmentations were excluded. Training that incorporated multiple lighting conditions alongside traditional augmentations further improved accuracy, suggesting that ray-traced lighting variation can be a valuable and complementary source of data augmentation for unsupervised contrastive learning.

\begin{table*}[ht!]
\caption{\textbf{Comparison of the ESS-MB trained on House100K with various augmentation settings} \hfill\break The column ``Augmentation'' indicates whether the pretext training uses the augmentation method from the original MoCo.}
\vskip 0.05in
\centering
\scalebox{0.8}{
    \begin{tabular}{l c p{0.2cm} l p{0.2cm} c c c}
    \toprule
        \multirow{2}{*}{Pretext Dataset} & \multirow{2}{*}{Augmentation} & & {Pretext Task} && \multicolumn{3}{c}{Downstream ImageNet Classification} \\
        &   & & Training loss $\downarrow$& & Training loss $\downarrow$& Test loss $\downarrow$&Test accuracy ($\%$) $\uparrow$\\
        \hline
        House100K& & &$4.08\pm 0.003$ & & $6.13\pm 0.229$ & $6.20\pm 0.18$ & $9.70\pm 0.16$\\
        House100KLighting& & & $4.07\pm 0.005$ & & $5.77\pm 0.314$ & $5.92\pm 0.35$ & $14.09\pm 0.25$\\
        House100K&$\checkmark$ & & $4.00\pm 0.005$ &  &$4.67\pm 0.002$&$4.75\pm 0.01$& $18.05\pm 0.04$\\
        House100KLighting& $\checkmark$ & & $4.03\pm 0.001$& &$4.49\pm 0.013$&$4.51\pm 0.01$&$\textbf{20.74}\pm 0.17$\\
    \bottomrule
    \end{tabular}}
    \label{tab:augmentation_result}
\end{table*}

\subsection*{ESS training improves localization}
To determine whether ESS-MB training is also superior in tasks related to spatial perception compared to MoCo, we developed two downstream tasks. The first task required the model to classify the specific room of a house based on a given view, while the second task required the model to predict the exact position and orientation of a provided view. For these evaluations, we compared ESS-MB with baseline models that had been pretrained on House100K. The room classification task was first conducted on images from the House14K dataset. However, the performance was very close to the ceiling so we created a more challenging variant where the lighting condition for each sample was varied randomly. As shown in Table~\ref{tab:cls_result}, the accuracy of ESS-MB on House14K, House14KLighting, and Apt14K surpassed the baseline model by $1.25\%$, $8.67\%$, and $14.99\%$, respectively. ESS-MB performs better in classifying the rooms in the environment than the baseline, especially when transferring to lighting conditions and environments not encountered during pretext training.

\begin{table*}[ht!]
\caption{\textbf{Comparison of the baseline and ESS-MB models trained on House100K on the room classification task for images from the House14K, House14KLighting and Apt14K datasets}\hfill\break }
\centering
\scalebox{0.8}{
    \begin{tabular}{llccc}
    \toprule
        Model& Dataset & Training loss $\downarrow$& Test loss $\downarrow$&Test accuracy ($\%$) $\uparrow$\\
        \hline
        Baseline & House14K & $0.19\pm 0.003$&$0.19\pm 0.004$& $98.10\pm0.08$\\
        ESS-MB & House14K&$0.08\pm 0.002$&$0.08\pm 0.002$&$99.35\pm 0.08$\\
        Baseline & House14KLighting &$0.88\pm 0.01$&$0.93\pm 0.01$&$78.70\pm 0.85$\\
        ESS-MB & House14KLighting&$0.47\pm 0.02$&$0.52\pm 0.03$&$87.37\pm 0.54$\\
        Baseline &Apt14K&$1.30\pm 0.04$&$1.30\pm 0.03$&$74.85\pm 0.26$\\
        ESS-MB &Apt14K&$0.65\pm 0.03$&$0.64\pm 0.03$&$89.84\pm 0.97$\\
    \bottomrule
    \end{tabular}}
    \label{tab:cls_result}
\end{table*}

In the spatial localization task, pretrained models are fine-tuned to estimate the position and rotation of the agent. As shown in Table~\ref{tab:reg_result}, ESS-MB consistently achieves lower losses compared to the baseline for both datasets. Specifically, ESS-MB predicts the position of images with an error of under 1 meter and 2 meters for House14K and Apt14K, respectively. ESS-MB training leads to better predictive accuracy in position by 0.15 meters in House14K and 0.51 meters in Apt14K.  
\begin{table*}[ht!]
\caption{\textbf{Comparison between the baseline and ESS-MB models trained on House100K for the spatial localization task}\hfill\break Position error represents the discrepancy in the predicted avatar position, denoted as $L_\text{pos.}$ in the text. Rotation error refers to the error in the predicted avatar rotation, denoted as $L_\text{rot.}$. Position drop and Rotation drop indicate the reduction in position error and rotation error from the start to the end of training, respectively.}
\vskip 0.05in
\centering
\scalebox{0.8}{
    \begin{tabular}{llcccccc}
    \toprule
        \multirow{2}{*}{Model} & Test & \multirow{2}{*}{Training loss $\downarrow$} & \multirow{2}{*}{Test loss $\downarrow$} & \multicolumn{2}{c}{Position (meter)} & \multicolumn{2}{c}{Rotation (degree)} \\
& dataset& & & Error $\downarrow$ & Drop $\uparrow$& Error $\downarrow$& Drop $\uparrow$\\
        \hline
        Baseline & House14K & $15.53\pm 0.16$&$15.24\pm 0.14$& $0.96\pm 0.01$&$2.12\pm 0.07$&$71.77\pm 0.34$&$34.25\pm 0.34$\\
        ESS-MB & House14K&$9.40\pm 0.19$&$9.21\pm 0.25$&$0.81\pm 0.01$&$1.61\pm 0.08$&$55.51\pm 0.83$&$50.37\pm 0.83$\\
        Baseline &Apt14K&$32.75\pm 0.29$&$33.35\pm 0.31$ & $2.35\pm 0.06$&$3.27\pm 0.09$&$100.11\pm 0.19$&$2.65\pm 0.20$\\
        ESS-MB &Apt14K&$26.96\pm 0.60$&$27.46\pm 0.68$&$1.84\pm 0.07$&$2.77\pm 0.03$&$93.06\pm 0.82$&$8.71\pm 0.77$\\
    \bottomrule
    \end{tabular}}
    \label{tab:reg_result}
\end{table*}

While both models exhibit notable rotation errors, ESS-MB outperforms the baseline in both tasks, with a superiority of 16.26 and 7.05 degrees for House14K and Apt14K, respectively.

\section*{Discussion}
\subsection*{Interpretation of the results}
Inspired by the processes of childhood learning, these results provide clear evidence that incorporating spatial context in environmental sampling significantly improves the effectiveness of contrastive learning compared to methods using an equivalent number of training epochs on the same dataset. Both rotation and position are important for defining whether a pair of views is similar enough. Moreover, the magnitude of the threshold for spatial similarity influences the learning outcome; excessively large thresholds might mislabel highly distinct views as positive pairs. Additionally, we discovered that resampling the same environment to acquire more images substantially boosts downstream accuracy, even if the images originate from identical rooms with the same furnishings and largely similar trajectories. Collectively these findings support the ability of visual learning algorithms to efficiently extract visual pattern information from a given environment, both by tracking the history of spatial information and denser reexploration of the same locations from slightly different positions and view angles as exemplified in Figure~\ref{fig:imgdif}A. 

Our approach is versatile and can be applied to contrastive learning with any dataset embedded with spatial history information. Furthermore, it holds the potential for adaptation to datasets rich in temporal sequence information, such as the Ego4D dataset.\cite{grauman2022ego4d} Here, temporal similarity could potentially replace spatial similarity. Moreover, our training experiments show that resampling the same views under different illuminants offers a source of augmentation (e.g., the trees in Figure~\ref{fig:imgdif}B) that complements traditional techniques, such as color manipulation. Additionally, the superior performance of ESS-MB in tasks like room classification and spatial localization demonstrates its ability to learn tasks associated with spatial perception, both within and across environments.

\subsection*{Implications of the study}
The long-term implications of this research span beyond developing general-purpose vision algorithms. It holds promise for embedded systems that need to learn in specific environments. The approach provides intelligent agents the ability to more rapidly learn generalizable visual understanding skills--achieved by tracking their location as they explore the environment and then performing either online or offline learning to improve performance for subsequent tasks. This would be helpful when a small drone dispatched to a remote location with unique lighting or other visual characteristics or a robotic explorer sent to a remote planet would require the acquisition of a new visual representation backbone while minimizing power consumption, making training efficiency a critical factor. Offline training could be performed using more efficient hardware connected to a power source and then the resultant backbones could be distributed to numerous drones for fine-tuning. The long-term impact of this work could therefore be significant for several sectors, including robotics, unmanned aerial vehicles, robot-assisted scientific exploration, disaster-relief operations, environmental surveillance in inaccessible locales, and planetary and space exploration. While our current focus is on the classification of static images, the potential exists for tasks that rely on contiguity between images such as action classification and navigation. Moreover, simulated environments offer a unique opportunity for designing augmentations that reflect the kind of changes that occur in the real world, potentially leading to more effective training for perceivers operating in real-life situations by including simulated datasets. Our positive results with lighting-based augmentation indicate that further exploration of this approach could be beneficial when the reflectance properties of surfaces and natural illuminants of an environment have been measured.

In addition to computer vision, spatial similarity training could also shed light on the invariance properties of human visual neurons that tolerate massive changes in an object's size, position, and rotation. This phenomenon could result from the natural temporal contiguity of visual input\cite{li2008unsupervised} or smooth changes in input features over time.\cite{wood2018development} This would be a potentially valuable method to simulate the development of visual neurons in simulations of biological visual systems. 

\subsection*{Limitations of the study}
A limitation of our study is the modest overall accuracy achieved in the downstream image classification task. This is a predictable outcome given such a small set of images used in training and the narrow scope of the environment from which they were collected--especially when compared with the diversity of ImageNet. However, this limitation mimics the real-world learning scenarios experienced by embodied agents, such as children, who learn a robust basis set of visual representations through exposure to restricted environments. The evident gap between our current downstream accuracy and human performance in image classification suggests significant opportunities for improvement and future development in training algorithms that exploit environmental context. Given the relatively small size of the datasets used in our framework,\begin{wrapfigure}[30]{r}{0.5\textwidth}
    \centering
\includegraphics[width=0.48\textwidth]{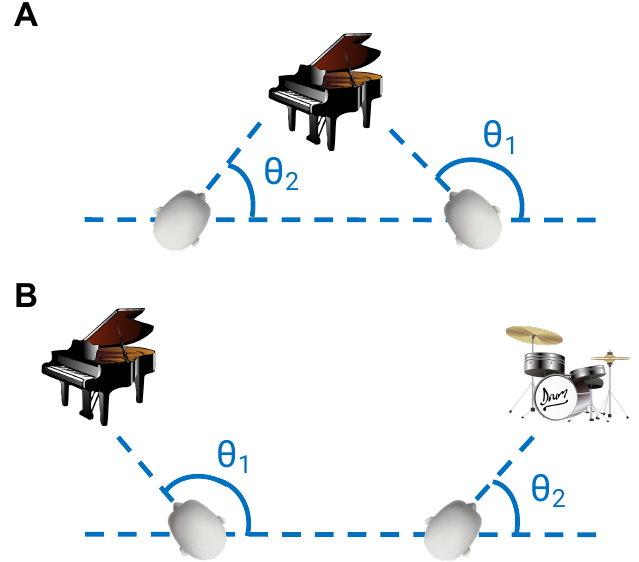}
    \caption{\textbf{Illustration of a more comprehensive approach to evaluating spatial similarity, which considers not only the distance and angle between two views but also the specific region of space being observed}\hfill\break
Even though the angular difference between the two views generated in (A) and (B), calculated as $ \lvert \theta_1-\theta_2 \lvert $, and the position difference are equivalent, the two views in (A) could be considered more similar due to their convergent perspective and shared focus on a specific region of space. In contrast, the views in (B) may be considered less similar due to their divergent perspective and lack of overlap in the region of space being observed.}
\label{fig:angles}
\end{wrapfigure} there is potential for rapid experimentation and iterative refinement of similar algorithms. In light of these findings, we encourage the computer vision community to explore ways to narrow the accuracy gap for such datasets. We have made all of our datasets available on an explanatory website (see Experimental procedures).

\subsection*{Future directions}
To further improve this approach, there are other aspects of ray-traced simulation that we have not explored. For example, the covariation of distance to an object and the camera's focal length alters the apparent size of different parts of the object based on their distance from the observer (e.g., the statue in Figure~\ref{fig:imgdif}C). This type of variation occurs naturally in real-world viewing conditions but cannot be accurately simulated through simple augmentations such as cropping and magnification.

Another opportunity for further improvement lies in refining the spatial similarity function used to identify positive pairs. Currently, our algorithm defines spatial context such that similarity between two data points decreases sharply with greater distance or differences in rotational angle separately. However, as shown in Figure~\ref{fig:angles}, even with an identical distance and rotation difference, the similarity between the two views can differ. There are ways to revise this function by incorporating information about pixel depth and objects. Depending on this function, our ESS-MW approach can be further explored and enhanced. For example by allowing samples with farther spatial separation to be flagged as slightly similar according to the presence or absence of intervening visual barriers or objects. This could be detected by tracking the variability in the visual input over time, such that passing through doorways or other barriers would cause dramatic shifts in the visual statistics, and therefore would down-weight the similarity of those samples. 

Besides, it is worth investigating the effect of increasing the number of images collected from a single environment on performance. Our analysis has shown that using 100K images as opposed to 14K images from the same house resulted in a significant improvement in downstream accuracy, even though both datasets contained images of the same rooms and the longer trajectory essentially covered the same views as the shorter one. It remains an open question how downstream accuracy would change with further increases in the density of image sampling from a given environment, and whether there is a ceiling to the accuracy attainable from a particular environment.

Lastly, interacting with the virtual platform TDW and conducting online learning is a further direction of exploration. An extra adaptive network can be trained to determine the movement direction and rotation of the avatar based on the avatar's field of view and historical information to maximize the information that can be gained from the environment. The current bottleneck is that the interaction between the avatar and the virtual environment cannot be processed in batches, which greatly affects the training speed. A possible alternative is to collect a dense dataset of images in advance, and then choose informative samples for training via the adaptive network.

\section*{Experimental procedures}

\subsection*{Resource availability}
\subsubsection*{Lead contact}
Request for information and resources used in this article should be addressed to Dr. Brad Wyble (bpw10@psu.edu).
\subsubsection*{Materials availability}
This study did not generate new unique reagents.
\subsubsection*{Data and code availability}
Our dataset is based on the high-fidelity 3D virtual environment, ThreeDWorld,\cite{gan2020threedworld} which can be downloaded at \url{https://www.threedworld.org}. Datasets used in this paper have been deposited at the OSF under DOI \url{10.17605/OSF.IO/W98GQ} and are publicly available as of the date of publication.\cite{lizhen2024dataset} All datasets are also available at \url{http://www.child-view.com}. We provide our two-stage dataset generation pipeline, along with the codes for conducting all the experiments and the pretraining and downstream checkpoints, at the OSF, 
under DOI \url{10.17605/OSF.IO/FT59Q} and are publicly available as of the date of publication.\cite{lizhen2024context} Any additional information required to reanalyze the data reported in this paper is available upon request.


\subsection*{Dataset generation process}
In the first stage, the selected environment was initialized with a set of predetermined objects and a non-kinematic default avatar was placed in a suitable location within the environment. All objects were given a mass of 10,000 to prevent movement due to avatar collisions. Using ThreeDWorld's interaction module, a user maneuvered the avatar, navigating its trajectory with functionalities like turning, advancing, retreating, and jumping--all triggered by specific keystrokes. The trajectory of the avatar, including the step numbers, positions, and rotations (represented by quaternions to avoid gimbal lock), was recorded as the agent traversed the house. The rotation of the avatar only changed in the horizontal (yaw) plane.

In the second stage, the same objects and avatar were placed in the environment and the skybox was configured either to its default setting or to one of the nine pre-selected skyboxes for the lighting augmentations. To ensure the quality of captured images, the resolution was set to $1024\times 1024$ and the field of view to $60$ degrees. Other parameters, such as render quality and shadow strength, were set to the default values in ThreeDWorld. The avatar retraced the earlier recorded trajectory, moving to the predetermined position and rotation at each step and capturing a $1024\times 1024$ RGB image. These images were resized to a $224\times224$ resolution using Python codes with antialiasing from the PIL library. This pipeline can also be used by researchers to generate datasets with customized settings. The environment initialization and avatar camera parameters are both adjustable. 

\begin{figure*}[ht!]
    \centering
    \includegraphics[width=0.98\textwidth]{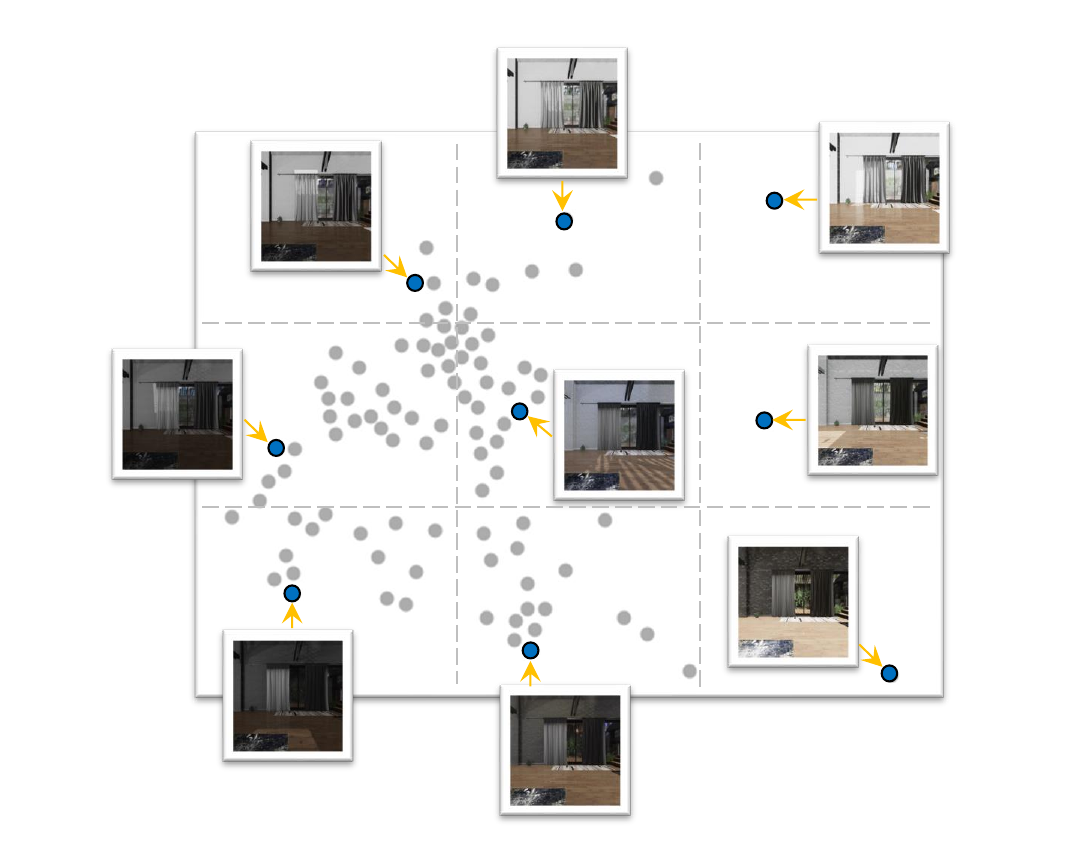}
    \caption{\label{fig:light}\textbf{Illustration of representative lighting conditions available in ThreeDWorld}\hfill\break A total of 95 lighting conditions are shown here, distributed according to a cluster analysis based on pixel values of three example images captured in the House environment using the t-SNE algorithm. From the total collection of skyboxes, nine were selected to cover this space. For each selected skybox, an example image, taken from an identical viewpoint within the house, is shown. From left to right and top to bottom, the skyboxes' names are as follows: Kiara\_1\_dawn, Ninomaru\_teien, Small\_hangar\_01, Venice\_sunrise, Blue\_grotto, Whipple\_creek\_gazebo, Mosaic\_tunnel, Royal\_esplanade, and Indoor\_pool.}
\end{figure*}

One important advantage of varying light sources in a ray-traced virtual environment is its capacity to more accurately emulate the real-world physics of light reflection, resulting in a richer variety than basic augmentation techniques that merely shift spectral distributions. The ThreeDWorld platform features 95 distinct skyboxes as environment lighting conditions. We controlled an avatar to capture three images from the living room, stairs, and bedroom, maintaining consistent position and rotation in the House environment for each of the 95 skyboxes. The t-distributed stochastic neighbor embedding (t-SNE)\cite{hinton2002stochastic} was then used to cluster concatenations of those three images simulated under each of the 95 skyboxes. To explore lighting augmentations, we selected nine skyboxes, drawn from a $3\times 3$ grid of the t-SNE plot (Figure~\ref{fig:light}). A sample image from the House environment for each chosen skybox is shown within the t-SNE plot. Every image in the House14k and House100K datasets was generated ten times, one with the default lighting condition of ThreeDWorld and also one for each of these nine skyboxes. The resulting datasets are titled House14KLighting and House100KLighting. For more details on the lighting models within ThreeDWorld, readers can refer to the primary reference.\cite{gan2020threedworld}

\subsection*{Implementation details of ESS-MB and ESS-MW}
\begin{figure*}[ht!]
    \centering 
    \includegraphics[width=0.95\textwidth]{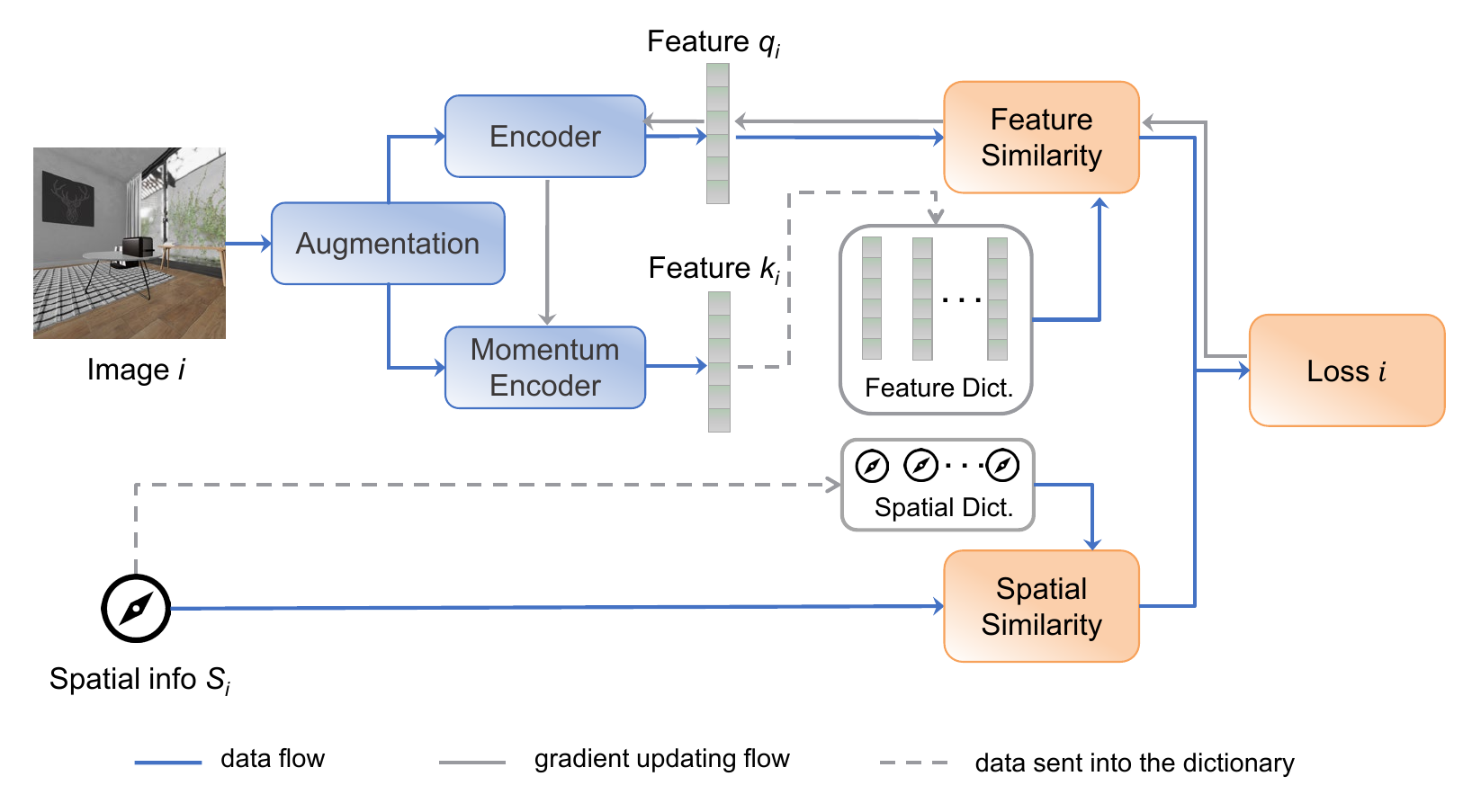}
    \caption{\textbf{The proposed ESS-MB approach}\hfill\break The learning algorithm compares a given image against the $N$ images in the dictionary, using their spatial position and rotation information to find positive pairs by comparing their relative spatial position and rotation values against a given threshold. The feature values of all images within the dictionary are then compared to compute the loss value relative to whether each image is part of a positive pair. This loss value is used to drive gradient descent as in the original MoCo formulation.  }   
\label{fig:pipelineB}
\end{figure*}

\begin{figure*}[ht]
    \centering 
    \includegraphics[width=0.95\textwidth]{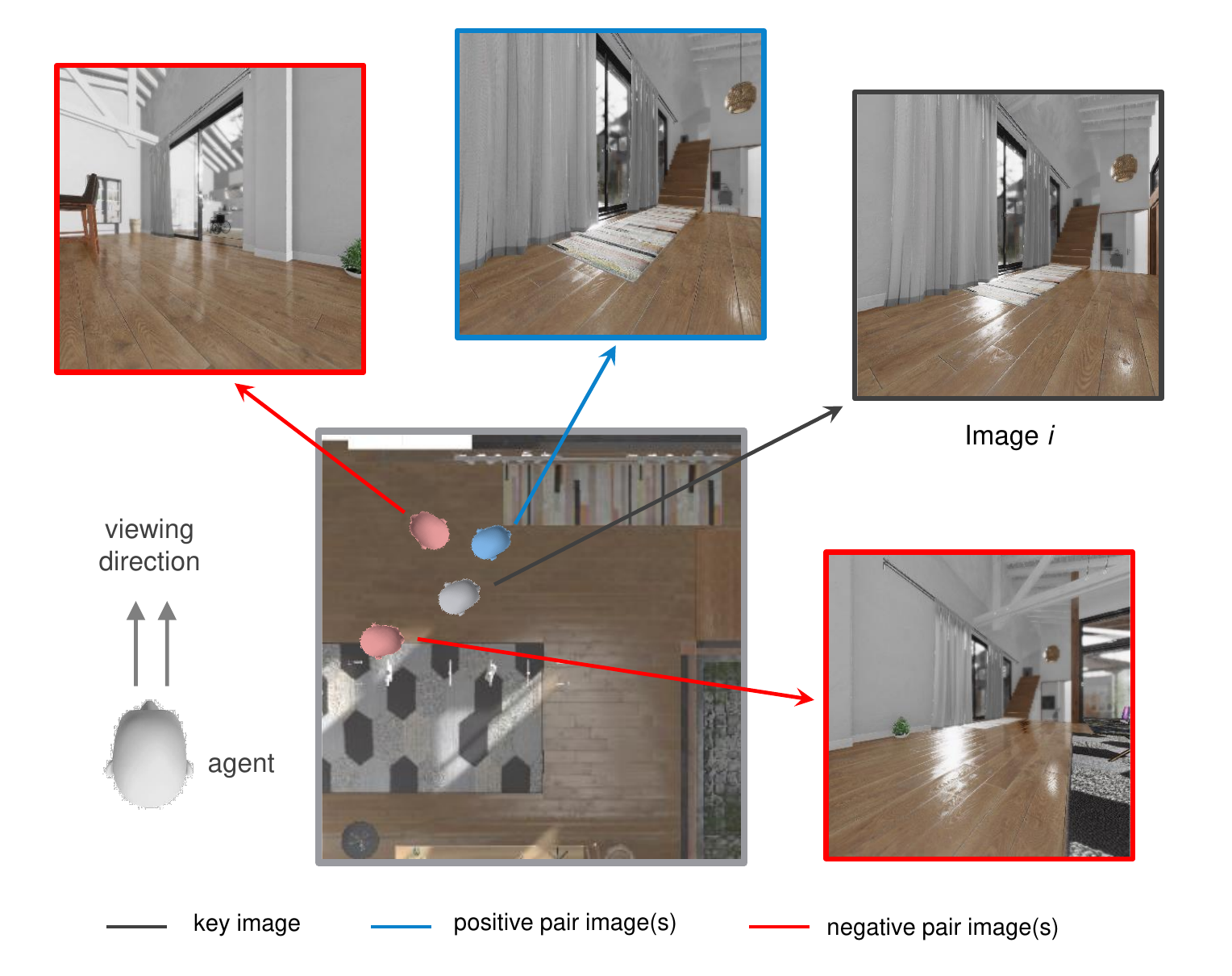}
    \caption{\textbf{The illustration of positive pair and negative pairs}\hfill\break Four different views of the agent in a room, based on the agent's location and viewing direction. Image $i$ and the blue image would be considered a positive pair. The rotation distance between the red image on the upper left and image $i$ is larger than the set threshold, hence they are considered a negative pair. Similarly, the position distance between the red image at the bottom right and image $i$ is larger than the set threshold, hence they are considered a negative pair.}   
\label{fig:pipelineA}
\end{figure*}

Our model is based on the MoCo V2 architecture,\cite{chen2020mocov2} implemented using Pytorch. ESS-MB randomly selects a fixed number of images from the dataset for each batch. As illustrated in Figure~\ref{fig:pipelineB}, each input image is transformed with randomly selected augmentation operations as in MoCo. Data augmentation techniques applied here included random cropping, Gaussian blur, horizontal flipping, color jittering, and grayscale conversion. Each transformed image $i$ is then encoded into two 128-dimensional vectors, called the query feature $q_i$ and key feature $k_i$, by the key encoder and the momentum encoder, respectively, which are both ResNet-50\cite{he2016deep} backbones that have different parameters. The $k_i$ feature is normalized and stored with its position and rotation information in a fixed-sized dictionary that records them as a queue. The dictionary size is set to 4,096 to accommodate the size of our comparatively small data set. The spatial information from which the image generating $q_i$ originated is compared to the spatial information linked to each feature in the dictionary. In contrast to conventional contrastive learning, our approach identifies a positive pair based on spatial similarity up to a certain threshold. The difference between positions $(x_i,y_i,z_i)$ and $(x_j,y_j,z_j)$ is calculated by the Euclidean distance:
\begin{equation}
\label{eq:pos}
\Delta_\text{pos.}=\left((x_i-x_j)^2+(y_i-y_j)^2+(z_i-z_j)^2\right)^{1/2}\;.
\end{equation}
The difference between rotations $r_i$ and $r_j$ is defined as: 
\begin{equation}
\label{eq:rotation}
\Delta_\text{rot.}=\min(|r_i-r_j|, 360-|r_i-r_j|)\;.
\end{equation}
The binary function to calculate the spatial similarity is defined by:
\begin{equation}
\label{eq:similarity}
f_{(\theta_\text{p},\theta_\text{r})}=
\begin{cases}
1& \text{if } \Delta_\text{pos.}<\theta_\text{p}\ \text{and}\ \Delta_\text{rot.}<\theta_\text{r}\;,\\
0& \text{otherwise},
\end{cases}
\end{equation}
where $\theta_\text{p}$ is the threshold of the position and $\theta_\text{r}$ is the threshold of the rotation. As illustrated in Figure~\ref{fig:pipelineA}, a pair of images with positional difference within a specified range (in meters) and rotational difference within a given range (in degrees) is considered a positive pair. Otherwise, they are labeled as a negative pair.

The loss function for image $i$ is then calculated as follows:
\begin{equation}
\label{eq:multi}
 L_i=-\frac{1}{|P(i)|}\sum\limits_{p\in P(i)} \log\frac{\exp(\text{sim}(q_i, k_{p})/\tau)}{\sum_{d\in D}\exp(\text{sim}(q_i, k_d)/\tau)}\;,
\end{equation}
where $\text{sim}(u,v)=u^Tv/(||u||||v||)$, represents the cosine similarity of two vectors, $P(i)$ represents the set of positive pairs with the key image $i$, and $\tau$ is the temperature parameter that controls how much attention is paid to difficult samples. The set $D$ represents the dictionary. 

This strategy makes use of the spatial information from the environment to define the positive pairs. As there are often multiple samples in the dictionary that fall within the spatial similarity threshold relative to the query image, we use ESS-MB, where MB indicates there are multiple positive pairs. This strategy ensures that we don't miss useful information or compromise training efficiency by only focusing on a single sample with high similarity to the query images. In ESS-MB, every sample within the spatial similarity threshold is treated as an equally valid positive pair during the calculation of loss, which is inspired by the supervised contrastive learning method.\cite{khosla2020supervised} This approach allows for a more comprehensive consideration of relevant samples, leading to improved performance compared to using just a single positive pair.

In contrastive learning, image similarity is binary, in that images are either identical or not. In the real world, the degree of similarity between two views is continuously changing based on changes in the position of the viewer. To capture this dynamism, in ESS-MW approach, we assign each positive pair of views $i$ and $j$ a weight $w_{i,j}$, which is defined as:
\begin{equation}
\label{eq:weight}
w_{i,j}=\frac{1}{\exp{\left(\alpha\left({\beta\Delta_\text{rot.}}+\Delta_\text{pos.}\right)\right)}}\;,
\end{equation}
where $\alpha$ controls the influence of spatial context differences and $\beta$ balances the relative importance of position and rotation in the weight calculation. The assigned weight increases in proportion to the similarity of the view pair. The loss function is defined as follows:
\begin{equation}
\label{eq:weighted}
 L_i=-\sum\limits_{p\in P(i)}\frac{w_{i,p}}{\sum\limits_{j\in P(i)} w_{i,j}} \log\frac{\exp(\text{sim}(q_i, k_{p})/\tau)}{\sum_{d\in D}\exp(\text{sim}(q_i, k_d)/\tau)}\;.
\end{equation}
As shown in Table\ref{tab:weightedparameter}, on House100K, the best downstream task performance can be achieved when $\alpha$ is $2$ and $\beta$ is $1/60$.

\begin{table*}[ht!]
\caption{\textbf{Comparison of ESS-MW with different hyperparameter values on House100K}\hfill\break }
\vskip 0.05in
\centering
\scalebox{0.8}{
    \begin{tabular}{l l l p{0.01cm} c p{0.01cm} c c c}
    \toprule
        \multicolumn{3}{c}{Training Stage} && {Pretext Task} && \multicolumn{3}{c}{Downstream ImageNet Classification} \\
         $\alpha$&$\beta$& Threshold &  &Training loss $\downarrow$ &  & Training loss $\downarrow$ & Test loss $\downarrow$ &Test accuracy $\uparrow$ \\
        \hline
         2&1/60&(0.8,12)& & $3.92\pm 0.003$ &  &$4.62\pm 0.001$&$4.67\pm 0.004$& $\textbf{18.39}\pm 0.082$\\
         2&1/30&(0.8,12)& & $3.91\pm 0.004$ &  &$4.62\pm 0.006$&$4.69\pm 0.001$& $18.13\pm 0.067$\\
         2&1/120&(0.8,12)& & $3.93\pm 0.001$ &  &$4.63\pm 0.011$&$4.70\pm 0.016$& $18.13\pm 0.134$\\
         1&1/60&(0.8,12)& & $3.97\pm 0.002$ &  &$4.66\pm 0.006$&$4.74\pm 0.011$& $17.69\pm 0.067$\\
         4&1/60&(0.8,12)& & $3.85\pm 0.002$ &  &$4.66\pm 0.007$&$4.70\pm 0.012$& $17.73\pm 0.102$\\
    \bottomrule
    \end{tabular}}
    \label{tab:weightedparameter}
\end{table*}

There is one implementation detail of ESS models that is worth noting. When comparing $q_i$ with features in the dictionary, if we first select the positive pairs from the dictionary before adding feature $k_i$, we call it the last-enqueue implementation. If the $k_i$ feature was added to the dictionary before selecting positive pairs, we define it as the first-enqueue implementation. In our implementation of the model trained on House14K and Apt14K, to prevent the model from selecting the other view of the same image with high probability (which would be similar to the original MoCo model), we used the last-enqueue implementation. However, last-enqueue sometimes led to a situation where there were no positive pairs in the dictionary, in which case a positive pair was generated by selecting the dictionary feature that had been generated by the image closest to image $i$ in the trajectory. This was a rare occurrence, on average happening with probability $0.03$ for House14K for the $0.5$ meters and $7.5$ degrees threshold and 0.02 for Apt14K with the 0.6 meters and 9 degrees threshold. For the model trained on House100K, the model often collapsed using last-enqueue, producing the same feature vectors for all inputs. To reduce this risk, we used the first-enqueue implementation on models trained with House 100K.

\subsection*{Pretext training}
The pretext task used 200 epochs and a batch size of 256. Due to the composition of our training set, we discovered that we could increase the learning rate from the initial 0.015 to 0.3 to increase accuracy and still have stable learning for both MoCo and our approach. Results from the House100K dataset, using the original learning rate, are provided in Supplemental Information. During training, the stop gradient method is applied to the momentum encoder. Only the main encoder parameters, $\theta_q$, are updated through backpropagation. The momentum encoder parameters, $\theta_k$, are updated by momentum updating:
$\theta_k\gets m\theta_k+(1-m)\theta_q$, where $m$ is the momentum coefficient. 

During the pretext training, all the training images were from virtual indoor settings, which markedly contrast with the samples in the downstream ImageNet classification task. We tried to improve the performance of the model by adding some ImageNetV2 images\cite{recht2019imagenet} into the training set. ImageNetV2 has 1,000 categories, with multiple images in each category that do not overlap with the standard ImageNet dataset used for the downstream task described below. Because ImageNetV2 doesn't include spatial information, for both the baseline and ESS-MB models, the only positive pair for any image is its augmented counterpart. For each training epoch, there are 102,197 images from House100K and an additional 10,000 images from ImageNetV2.

\subsection*{Evaluation of the learned representations}

\subsubsection*{The accuracy of the pretext task} 
The baseline model only considered one positive pair. The accuracy computation for the pretext training is different from ESS-MB which has multiple positive pairs and the two cannot be directly compared. In the baseline model, accuracy was calculated by determining if the pair with the highest cosine similarity was the pre-defined positive pair. For ESS-MB, accuracy was computed by applying the $sigmoid$ function to each cosine similarity score. If the result was greater than the threshold of 0.95, the pair was predicted to be positive; otherwise, it was predicted to be negative. The predicted result was then compared to the pre-defined positive pair according to the positions and rotation to calculate the accuracy. The pretraining accuracies of the main experiments from Table~\ref{tab:main_result} are shown in Table~\ref{tab:pre_acc}.

\subsubsection*{ImageNet classification task}
\begin{wraptable}[17]{r}{0.5\textwidth}
\caption{\textbf{Pretext training accuracies}}
\vskip 0.05in
\centering
    \scalebox{0.8}{
    \begin{tabular}{l l l c}
    \toprule
        Pretext dataset & Model & Threshold & Accuracy (\%) \\
        \hline
         \multirow{6}{*}{House100K}& Baseline&N/A &$82.11\pm 0.24$\\
         &ESS-MB&(0.4,6) &$99.57\pm 0.00$\\
         &ESS-MB&(0.8,12) &$99.37\pm 0.01$\\
         &ESS-MB&(1.6,24)&$98.89\pm 0.00$\\
         &ESS-MB&(0.8,N/A) &$98.77\pm 0.00$\\
         &ESS-MB&(N/A,12) &$97.66\pm 0.00$\\
         \hline
         \multirow{2}{*}{House14KLong}& Baseline&N/A &$54.92\pm 0.14$\\
         & ESS-MB &(0.5,7.5) &$99.59\pm 0.00$\\
         \hline
         \multirow{4}{*}{House14K}& Baseline&N/A &$18.43\pm 0.09$\\
         &ESS-MB &(0.25,3.75)& $92.11\pm 0.79$\\
         &ESS-MB &(0.5,7.5) &$93.50\pm 0.16$\\
         &ESS-MB &(1.0,15) & $92.55\pm 0.37$\\
         \hline
         \multirow{4}{*}{Apt14K}& Baseline&N/A &$17.57\pm 0.10$\\
         &ESS-MB &(0.3,4.5)& $91.27\pm 0.63$\\
         &ESS-MB &(0.6,9) &$91.87\pm 0.05$\\
         &ESS-MB &(1.2,18) &$91.54\pm 0.77$\\
    \bottomrule
    \end{tabular}}
    \label{tab:pre_acc}
\end{wraptable}
To evaluate the quality of the learned representations, as in MoCo, we added a linear classifier on top of the fixed backbone architecture and trained only the last added layer for 50 epochs of the ImageNet.

\subsubsection*{Room classification task}
In this task, we trained a linear classifier to label a given image according to what room it had been generated in using the features from each pretrained model. Each downstream model was trained for 20 epochs. The House environment includes 8 rooms, while the Apt environment consists of 9 rooms. Each image is labeled with a number, ranging from 0 to 7 (for House) or up to 8 (for Apt), to represent the room where it was captured. The boundaries and illustrations of each room are included in Supplemental Information. In each dataset, $80\%$ of images were used for the training and the remaining $20\%$ for testing. In the House14KLighting dataset, Mosaic\_tunnel and Venice\_sunrise lighting conditions were only applied to the test data. Meanwhile, each training image was randomly assigned one of the other seven lighting conditions.

\subsubsection*{Spatial localization task}
We added a single-layer neural network with four output nodes at the end of the pre-trained model. The training utilized $80\%$ of the images from each dataset, setting aside the remaining $20\%$ for testing purposes.  
In the spatial localization task, pretrained models are fine-tuned to estimate the position ($x_\text{p}$, $y_\text{p}$, $z_\text{p}$) and rotation $r_p$ of each image from the House14K and Apt14K datasets. The loss function, denoted as $L$, is defined by 
\begin{eqnarray}
L_\text{pos.}&=&\left((x_\text{p}-x)^2+(y_\text{p}-y)^2+(z_\text{p}-z)^2\right)^{1/2}\;, \\
L_\text{rot.}&=&\min(|r_\text{p}-r|,360-|r_\text{p}-r|)\;, \\
L&=&L_\text{pos.}^2+\alpha L_\text{rot.}^2\;,
\end{eqnarray}
where $\alpha$ is a hyperparameter for adjusting the ratio of $L_\text{pos.}$ and $L_\text{rot.}$. Here, we set $\alpha$ to $1/360$ to ensure both terms start with comparable magnitudes. 

\subsection*{Hyperparameter sensitivity analysis}
Several hyperparameters play a role in pretraining and may indirectly affect the downstream performance. We trained a series of ESS-MB models on the House100K dataset, adhering to the pipeline described in Contrastive Learning Models for Experimental Procedures section. We varied the batch sizes, temperature parameters, thresholds, and dictionary sizes of the original ESS-MB model, either doubling or halving them individually. Additionally, we tested the use of both default and multi-skybox settings.

\begin{figure*}[ht!]
    \centering
    \includegraphics[width=0.7\textwidth]{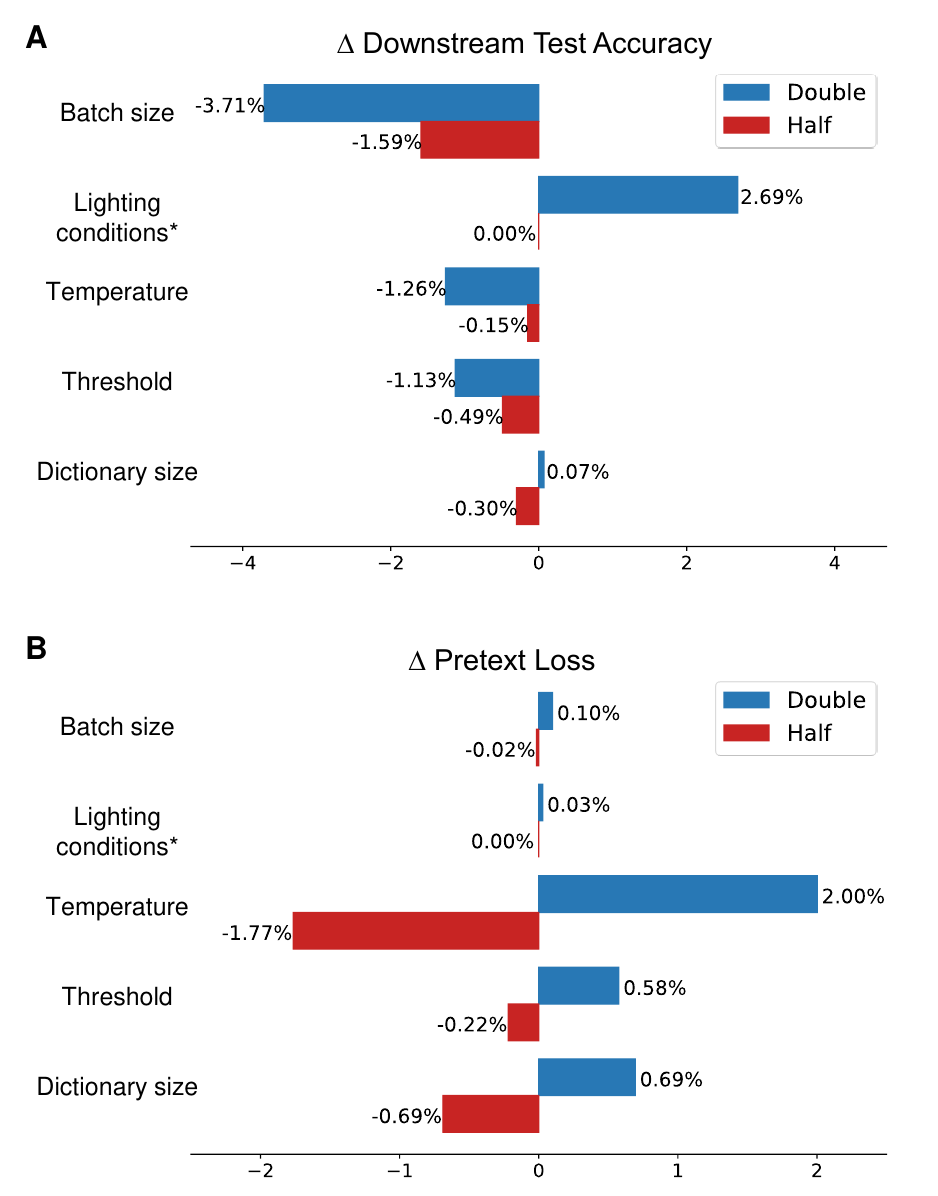}
    \caption{\textbf{Impact of varied pre-training hyperparameters on downstream test accuracy and pretext loss }\hfill\break 
    (A) The effect of hyperparameter variations on downstream test accuracy. (B) The effect of hyperparameter variations on pretext loss during pretraining. Regarding batch sizes, temperature parameters, thresholds, and dictionary sizes, blue indicates that the parameter value has been doubled, while red indicates that it has been halved. $^*$ For lighting conditions, blue represents the default setting, whereas red represents the use of multiple skyboxes. }
\label{fig:tornado}
\end{figure*}

The results, as shown in Figure~\ref{fig:tornado}, indicate that modifications in batch sizes, lighting conditions, temperature parameters, thresholds, and dictionary sizes during pretraining have impacts on downstream accuracy. When the batch size was doubled or halved, the downstream test accuracy decreased by $3.71\%$ and $1.59\%$, respectively. Doubling the batch size caused the model not to converge well; halving the batch size, despite slightly reducing the pretext loss, limited the model's generalizability beyond House100K. Using multi-skybox augmentation enhanced the model's ability to generalize to other datasets. The temperature parameter,  which directly influences the loss function and determines the model's focus on harder samples during training, also showed a significant impact: doubling or halving it led to a decrease in accuracy by 1.26\% and 0.15\%, respectively. As discussed in the Experimental Procedures section, the threshold, which dictates the similarity criterion for positive pairs and the number of such pairs in the dictionary, also affects the results. In contrast, the dictionary size, determining the number of pairs to compare with the key sample, both positive and negative, for comparison with a key sample, had a more limited influence on the downstream task, as the similarity of positive pairs was already fixed.

\section*{Acknowledgments}
This material is based upon work supported in part by the National Science Foundation under Grant No. BCS-2216127. The Institute for Computational and Data Science of Penn State provided support to initiate this research. This work used cluster computers at the Pittsburgh Supercomputer Center through an allocation from the Advanced Cyberinfrastructure Coordination Ecosystem: Services \& Support (ACCESS) program, which is supported by NSF Grants Nos. 2138259, 2138286, 2138307, 2137603, and 2138296. This work also used the Extreme Science and Engineering Discovery Environment, which was supported by NSF Grant No. 1548562. The authors wish to express their gratitude to Molly Huang, Sitao Zhang, Yimu Pan, Dheeraj Varghese, and Hyungsuk Tak for their valuable contributions, insights, and discussions. In addition, we extend our appreciation to the ThreeDWorld team and, in particular, Jeremy Schwartz for their technical support and access. We thank the anonymous reviewers and the editor Wanying Wang for their constructive feedback.

\section*{Author contributions}
Conceptualization, B.W.; methodology, L.Z., B.W., and J.Z.W.; software, validation, and investigation, L.Z.; writing – original draft, L.Z. and B.W.; formal analysis, L.Z.; data curation, W.L. and B.W.; writing – review \& editing, J.Z.W. and B.W.; visualization, L.Z., W.L., J.Z.W., and B.W.; resources, supervision, project administration, and funding acquisition, B.W. and J.Z.W.

\section*{Declaration of interests}
The authors declare no competing interests.

\section*{Declaration of AI and AI-Assisted Technologies in the Writing Process}
During the preparation of this work, the authors used ChatGPT in order
to improve the readability and language of the work by fine-tuning some of the grammar. After using this
tool, the authors reviewed and edited the content as needed and take
full responsibility for the content of the publication.

\bibliography{references}



\section*{Supplementary material}

# Supplemental Experimental Procedures

**Room boundaries in two environments**

The Archviz House ('House') consists of eight distinct rooms, with example images depicted in Figure S1. These rooms are the kitchen, lower hall, lower bedroom, outer deck, upstairs piano room, bathroom, and upstairs bedroom. Images are classified based on their position inside the defined boundaries of each room. Certain images, such as those captured on the stairs, do not fall within the boundaries of any room and are therefore excluded from the evaluation. In House14K, 12,127 of 14,766 images are labeled. In House100K, 83,300 of 102,197 images are labeled. The boundaries and the number of samples for each room are listed in Table S1.

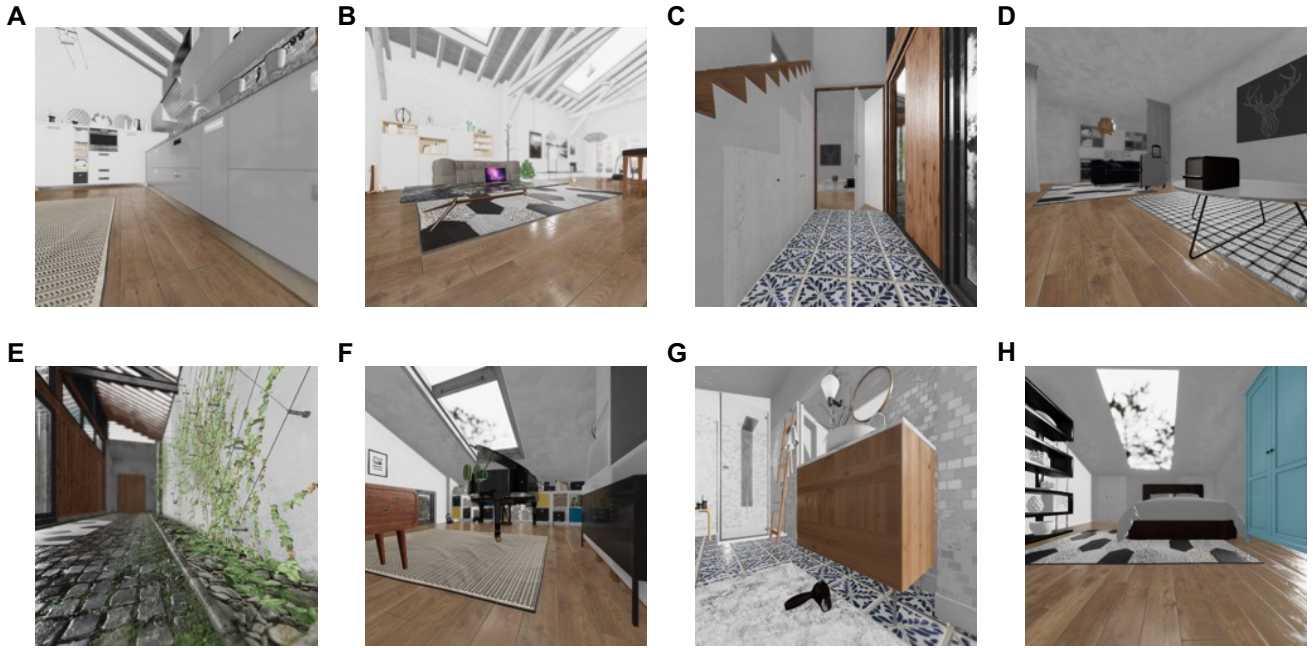

**Figure S1: Eight rooms in the House environment**
(A) Kitchen. (B) Living room. (C) Lower hall. (D) Lower bedroom. (E) Outer deck. (F) Upstairs piano room. (G) Bathroom. (H) Upstairs bedroom.

| Room name | $x$-axis boundary | $y$-axis boundary | Height boundary | House14K images | House100K images |
|---|---|---|---|---|---|
| kitchen | (-17.00, -11.83) | (-1.48, 1.80) | (1.4, 3.5) | 2079 | 7535 |
| living room | (-17.00, -7.60) | (-7.00, -1.48) | (1.4, 5.2) | 3225 | 32412 |
| lower hall | (-6.30, -4.10) | (-4.30, -3.10) | (0.6, 3.5) | 461 | 3490 |
| lower bedroom | (-3.43, 0.05) | (-4.30, 1.60) | (0.6, 3.5) | 1506 | 10155 |
| outer deck | (-7.10, 6.00) | (-7.40, -4.75) | (0.4, 5.2) | 2491 | 6721 |
| upstairs piano room | (-3.20, -0.25) | (-4.30, -3.00) | (3.5, 5.2) | 925 | 9882 |
| bathroom | (-3.20, -0.25) | (-4.30, -3.00) | (3.5, 5.2) | 419 | 2853 |
| upstairs bedroom | ( 0.00, 4.10) | (-4.30, 1.37) | (3.5, 5.2) | 1021 | 10252 |

Table S1: **The boundaries of the eight rooms in the House environment and the number of samples for each room**
The coordinate ranges are measured in the ThreeDWorld virtual environment. "House14K images" and "House100K images" means the number of images in each category for House14K and House100K respectively.

The Apartment ('Apt') layout consists of nine rooms, arranged in two rows. Rooms in the upper row in the floor plan are marked as rooms 0 to 4, from left to right. Rooms 5 to 8 are in the lower row in the same left-to-right sequence. The items placed in each room are carefully designed. For instance, entertainment facilities are filled in room 0. Rooms 3, 4, 6, and 7 serve as living rooms, each having a distinct style. Room 8 is a kitchen. In

Apt14K, 9,855 of 14,487 images are labeled. Sample images and the specified boundaries for each room are shown in Figure S2 and Table S2, respectively.

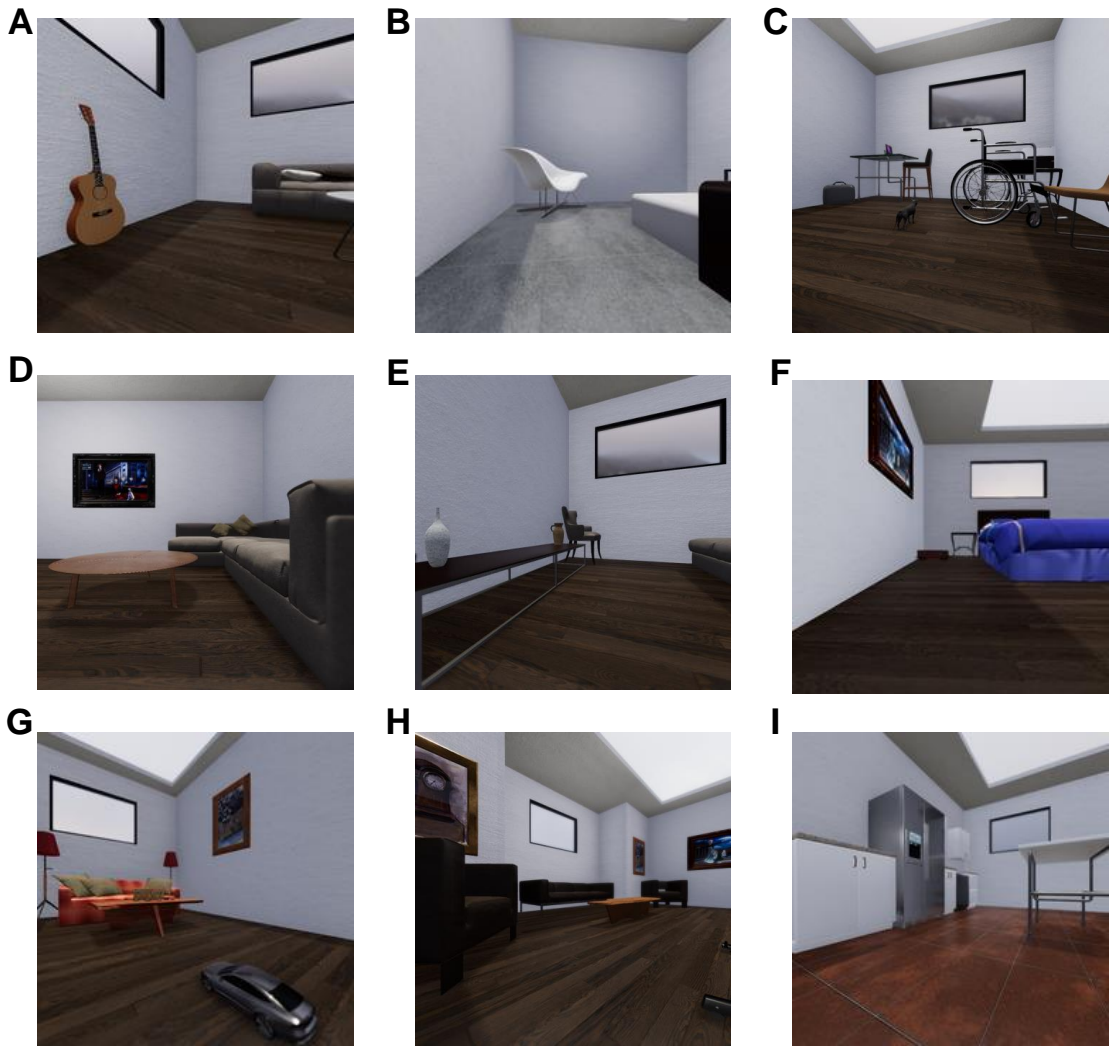

**Figure S2: Nine rooms in the Apt environment**
Letters A-I represent the rooms 0 through 8 in order.

| Room label | $x$-axis boundary | $y$-axis boundary | Images |
|---|---|---|---|
| 0 | (-10.6, -7.2) | ( 1.45, 4.80) | 680 |
| 1 | (-7.2, -3.1) | ( 1.45, 4.80) | 659 |
| 2 | (-3.1, 0.8) | ( 1.45, 4.80) | 1312 |
| 3 | ( 0.8, 6.8) | ( 1.45, 4.80) | 1295 |
| 4 | ( 6.8, 9.8) | ( 1.45, 4.80) | 749 |
| 5 | (-10.6, -6,2) | (-5.70, -0.18) | 1333 |
| 6 | (-6.2, -3.1) | (-5.70, -0.18) | 794 |
| 7 | (-3.1, 3.3) | (-5.70, -0.18) | 1269 |
| 8 | ( 3.3, 9.8) | (-5.70, -0.18) | 1764 |

Table S2: **The boundaries of the nine rooms in the Apt environment and the number of samples for each room**
The coordinate ranges are measured in the ThreeDWorld virtual environment.

**Learning rate comparison**

In our pretext training, we adopted a learning rate of 0.3 instead of the suggested $0.015$ from MoCo V2. This adjustment was made based on its improved overall accuracy on our datasets. Table S3 shows the results of MoCo V2 with different learning rates when trained on the House100K dataset.

| Learning rate | Pretext Task training loss | Pretext Task accuracy | Downstream ImageNet Classification Training loss | Downstream ImageNet Classification Test accuracy (%) |
|---|---|---|---|---|
| 0.015 | 3.73 | 73.45 | 209.25 | 7.61 |
| 0.3 | 4.43 | 82.11 | 4.71 | 17.36 |

Table S3: **Comparison results of MoCo V2 with two different learning rates trained on the House100K dataset**

**Details of the implementation of ESS-MB on other models**

SimCLR[S1] is a popular contrastive learning model in which the positive pair of an augmented view is itself. Negative pairs are other augmented samples from the same batch. Our ESS-MB on SimCLR found positive samples from the same batch according to spatial information. All parameters were the same as those in SimCLR. We ran the experiment on a single GPU for 200 epochs as suggested by the code.

DCL[S2] removes the positive pairs' effect on the denominator of InfoNCE loss. We implemented the updated loss function based on our original ESS-MB model for both DCL and ESS-MB with DCL. All the parameters were the same as the ESS-MB on MoCo.

CLSA[S3] categorizes augmentation operations into 'strong' and 'weak' and tries to align the feature distance distribution of views derived from these two augmentation types when finding the positive pairs from the weak augmented samples simultaneously. CLSA inherits the structure of MoCo. Based on the implementation of CLSA, our approach found positive pairs of an augmented view from the dictionary. In our experiment, we kept the hyperparameters of CLSA but modified the dictionary size and learning rate to match our original ESS-MB.

NNCLR[S4] computes similarity according to the proximity within a latent space generated by the encoder to contrastively learn representations from unlabeled images. The Lightly package was used to run NNCLR simulations with the ResNet-18 backbone. To ensure a fair comparison, we switched the backbone of ESS-MB on MoCo to Resnet-18 and trained both models for 200 epochs in the pretraining phase.

MoCo V3[S5] applied the contrastive learning structure to the Vision Transformer[S6] backbone. The dictionary size is set to 4096 to align with the threshold of ESS-MB approach. We run both models for 200 epochs in the pretraining with 256 batch size.

**Assessing the clustering of learned features**

After training on the datasets, we applied t-SNE[S7] on the features for a subset of images from the corresponding datasets to determine whether the training produced clustering of features from spatially proximal images. We randomly selected approximately 10,000 images that were inside the room boundaries and took the features from the fully trained ResNet model as input to the t-SNE. In the resulting t-SNE space, each image's features were labeled with a number (and a corresponding color), indicating the room of its origin.

The t-SNE visualizations generated for both the baseline and ESS-MB models trained on House and Apt environments are shown in Figures S3 and S4, respectively. Furthermore, we used the Silhouette Coefficient,[S8] Calinski-Harabasz index,[S9] and Davies-Bouldin index[S10] as metrics to assess the clustering quality of the t-SNE outputs. These results are shown in Table S4. Both results show a model trained on the more extensive House100K dataset exhibited a stronger capability in distinguishing features generated from different rooms compared to the one trained on the smaller House14K dataset. For all the models trained on three datasets, ESS-MB exhibited reduced clustering relative to the baseline model. This might be attributed to the ESS-MB training approach, which tends to group features of spatially proximal images together, regardless of room boundaries. In contrast, the baseline MoCo model relies only on instance discrimination. As a result, cluster boundaries for nearby locations in adjacent rooms would not be as distinct with ESS-MB.

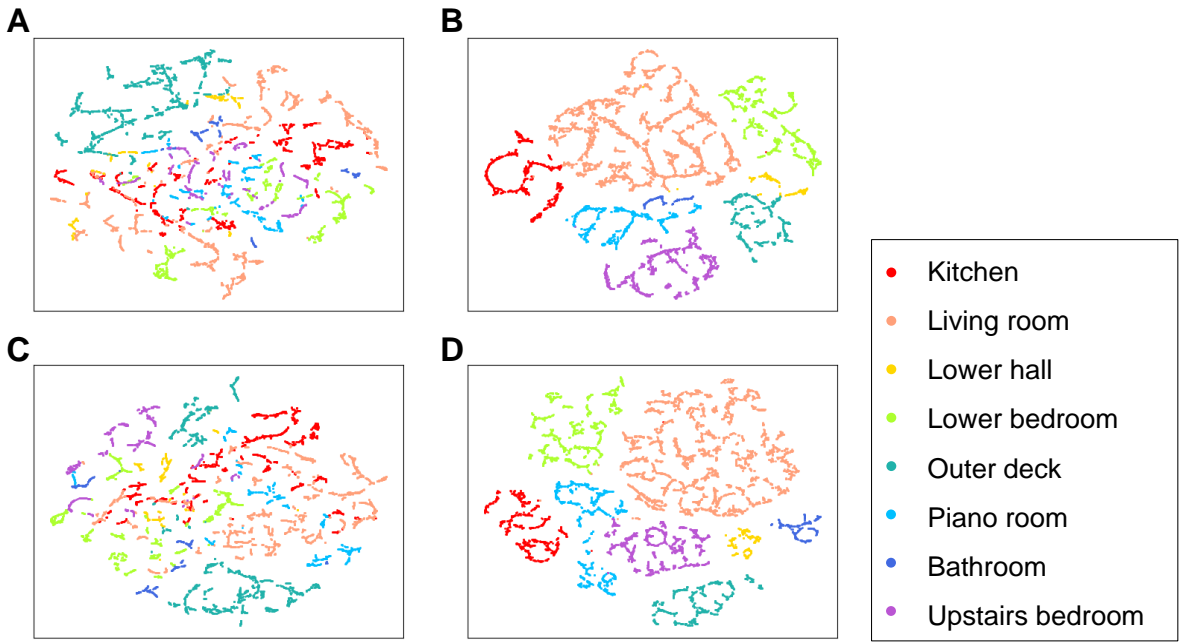

**Figure S3: The t-SNE results of the learned features in the House environment**
(A) ESS-MB on House14K. (B) ESS-MB on House100K. (C) Baseline on House14K. (D) Baseline on House100K.

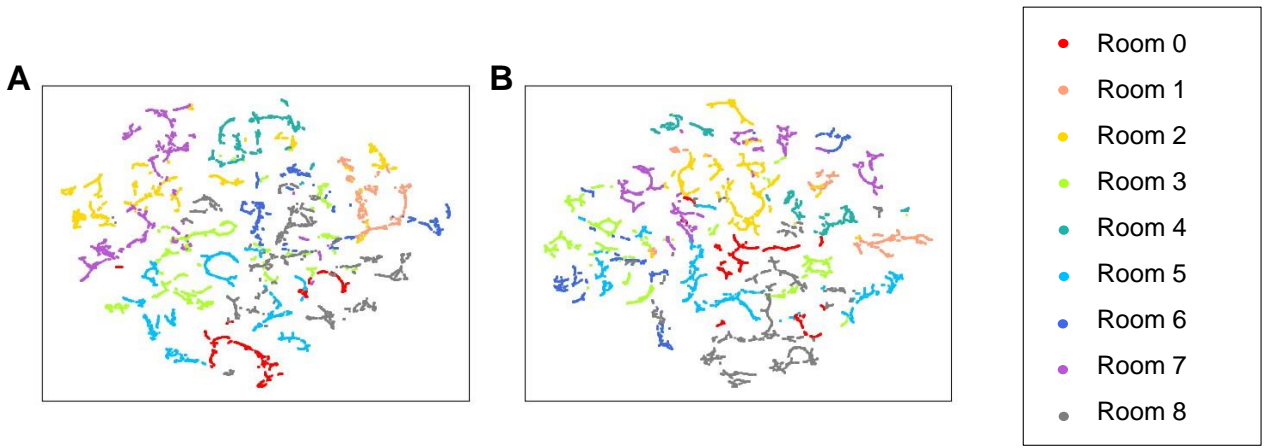

**Figure S4: The t-SNE results of the learned features in the Apt environment**
(A) ESS-MB on Apt14K. (B) Baseline on Apt14K.

| Pretext dataset | Model | Silhouette ↑ | CH index ↑ | DB index ↓ |
|---|---|---|---|---|
| House100K | Baseline | 0.2394 | 4538.16 | 0.8552 |
| House100K | ESS-MB | 0.1437 | 2393.92 | 1.8488 |
| House14K | Baseline | -0.0548 | 1187.78 | 5.5367 |
| House14K | ESS-MB | 0.0972 | 1549.44 | 9.8386 |

Table S4: **Evaluation of the learned features**
CH and DB stand for Calinski-Harabasz and Davies-Bouldin indices, respectively. An upward arrow indicates that a higher value for the respective index denotes more effective clustering, while a downward arrow implies the reverse.



\end{document}